%% file: main.tex
\documentclass[conference]{IEEEtran}
\usepackage{times}

\usepackage[numbers,sort&compress]{natbib}  
\usepackage{multicol}
\usepackage[bookmarks=true]{hyperref}

\usepackage{algorithm}
\usepackage{algpseudocode}
\usepackage{amsmath}
\usepackage{amssymb}

\usepackage{booktabs}
\usepackage{graphicx}
\usepackage{microtype}
\usepackage{xcolor}
\usepackage{xparse}

\input{x-notation}

\def\Snospace~{\S{}}

\definecolor[named]{xDarkGray}{HTML}{404040}

\definecolor[named]{TODORed}{HTML}{FF0000}
\definecolor[named]{TLDRViolet}{HTML}{800080}
\definecolor[named]{MissingCyan}{HTML}{47D4FF}
\definecolor[named]{PendingSaffron}{HTML}{F9C22E}
\definecolor[named]{SkeletonGray}{HTML}{808080}

\definecolor[named]{cBlue}{HTML}{18647E}
\definecolor[named]{cOrange}{HTML}{FF9B00}  
\definecolor[named]{cPink}{HTML}{F26DF9}
\definecolor[named]{cYellow}{HTML}{F0C808}
\definecolor[named]{cGreen}{HTML}{008000}

\def \papermode{draft}  

\def \draftmode{draft}

\ifx \papermode \draftmode
    \newcommand{\todo}[1]{{\color{TODORed}#1}}
    \newcommand{\TODO}[1]{{\color{TODORed}\textbf{[TODO]: #1}}}
    
    \newcommand{\tldr}[1]{\medskip \noindent {\color{TLDRViolet}{[TL;DR] :: #1}}}
    
    \newcommand{\needcite}{{\color{MissingCyan}\textbf{[NEED CITE]}}}

    \NewDocumentCommand{\makecomment}{m m m o}{%
      \noindent{\color{#2}[\textbf{#1}]: #3}%
      \IfValueT{#4}{ {\color{SkeletonGray}\textit{#4}}}%
    }

\else
    \newcommand[\todo][1]{}
    \newcommand[\TODO][1]{}
    \newcommand{\tldr}[1]{}

    \newcommand{\needcite}[1]{}
    \newcommand{\warning}[1]{}
    \newcommand{\maybe}[1]{}
    \NewDocumentCommand{\makecomment}{m m m o}{}
\fi

\NewDocumentCommand{\zhenyang}{m o}{\makecomment{ZC}{cYellow}{#1}[#2]}

\NewDocumentCommand{\licho}{m o}{\makecomment{LW}{cPink}{#1}[#2]}

\NewDocumentCommand{\atian}{m o}{\makecomment{AT}{cGreen}{#1}[#2]}

\NewDocumentCommand{\sidd}{m o}{\makecomment{SK}{cOrange}{#1}[#2]}

\NewDocumentCommand{\danfei}{m o}{\makecomment{DX}{cBlue}{#1}[#2]}

\begin{document}

\title{ReSteer: Quantifying and Refining the \\ Steerability of Multitask Robot Policies}

\author{Zhenyang Chen, Alan Tian*, Liquan Wang*, Benjamin Joffe, Yingyan Celine Lin, \\ 
Yuxiao Chen, Siddharth Karamcheti‡, Danfei Xu‡ \\
\normalsize Georgia Institute of Technology, *Equal contribution, ‡Equal advising \\
\textbf{\textcolor{cyan}{\href{https://resteer-vla.github.io/}{resteer-vla.github.io}}
}}
\maketitle

\begin{abstract}
\input{sections/00_abstract}
\end{abstract}
\IEEEpeerreviewmaketitle

\section{Introduction}
\label{sec:introduction}
\input{sections/01_introduction}

\section{Related Work}
\label{sec:related-work}

\input{sections/02_related-work}

\section{Quantifying Steerability}
\label{sec:quantifying-steerability}
\input{sections/03_quantifying-steerability}

\section{ReSteer: A Data-Centric Framework for Refining Steerability}
\label{sec:resteer-method}
\input{sections/04_resteer-method}

\section{Simulation Experiments}
\label{sec:sim-experiments}
\input{sections/05_sim-experiments}

\section{Real-World Experiments}
\label{sec:real-experiments}
\input{sections/06_real-experiments}

\section{Discussion \& Conclusion}
\label{sec:discussion}
\input{sections/07_discussion}

\clearpage                      
\bibliographystyle{unsrtnat}
\bibliography{references}

\newpage
\onecolumn
\appendices

\input{sections-appendix/x_overview}

\end{document}

%% file: x-notation.tex
\newcommand{\defeq}{\triangleq}
\newcommand{\given}{\,\big|\,}

\newcommand{\StateSpace}{\mathcal{S}}                 
\newcommand{\ActionSpace}{\mathcal{A}}                
\newcommand{\TaskIndexSet}{\mathcal{K}}               
\newcommand{\TransitionKernel}{\mathcal{P}}           

\newcommand{\state}{s}                                
\newcommand{\traj}{\tau}                              
\newcommand{\policy}{\pi}                             

\newcommand{\LangInstr}[1]{l_{#1}}                    

\newcommand{\TrajDistPi}[2]{p^{#1}_{#2}(\traj \given \state)}   

\newcommand{\DemoSet}[1]{D_{#1}}                      
\newcommand{\NumDemos}[1]{N_{#1}}                     

\newcommand{\StateVisitSet}[1]{\mathcal{S}_{#1}}      

\newcommand{\StateInterSet}[2]{\mathcal{S}^{\mathrm{inter}}_{#1,#2}}    

\newcommand{\StateUnionSet}[2]{\mathcal{S}^{\mathrm{union}}_{#1,#2}}    



%% file: sections/00_abstract.tex
Despite strong multi-task pretraining, existing policies often exhibit poor task steerability. For example, a robot may fail to respond to a new instruction ``put the bowl in the sink" when moving towards the oven, executing ``close the oven", even though it can complete both tasks when executed separately. We propose ReSteer, a framework to quantify and improve task steerability in multitask robot policies. We conduct an exhaustive evaluation of state-of-the-art policies, revealing a common lack of steerability. We find that steerability is associated with limited overlap among training task trajectory distributions, and introduce a proxy metric to measure this overlap from policy behavior. Building on this insight, ReSteer improves steerability via three components: (i) a steerability estimator that identifies low-steerability states without full-rollout evaluation, (ii) a steerable data generator that synthesizes motion segments from these states, and (iii) a self-refinement pipeline that improves policy steerability using the generated data. In simulation on LIBERO, ReSteer improves steerability by 11\% over 18k rollouts. In real-world experiments, we show that improved steerability is critical for interactive use, enabling users to instruct robots to perform any task at any time. We hope this work motivates further study on quantifying steerability and data collection strategies for large robot policies.

%% file: sections/01_introduction.tex







Language-conditioned multitask robot policies such as Vision-Language-Action (VLA) models~\citep{intelligence2025pi05visionlanguageactionmodelopenworld, kim2024openvlaopensourcevisionlanguageactionmodel, trilbmteam2025carefulexaminationlargebehavior, lee2025molmoact, brohan2023rt1, nvidia2025gr00tn1openfoundation, smolvla2025} have emerged as a promising paradigm for robotic control, unifying language-specified goals and visual understanding. 
These policies use language instructions as an intuitive and flexible task representation, leveraging pretrained multimodal backbones~\citep{radford2021learningtransferablevisualmodels} to encode grounded priors about the physical world. 
While these models have demonstrated immense capability, fitting dozens of tasks~\citep{trilbmteam2025carefulexaminationlargebehavior, brohan2023rt1}, generalizing across new embodiments~\citep{intelligence2025pi05visionlanguageactionmodelopenworld, lee2025molmoact}, and unlocking different types of embodied reasoning~\citep{lee2025molmoact, zawalski2024ecot}, a key failure mode remains: \textit{steerability}.

\begin{figure}[t]
    \centering
    \includegraphics[width=1.0\linewidth]{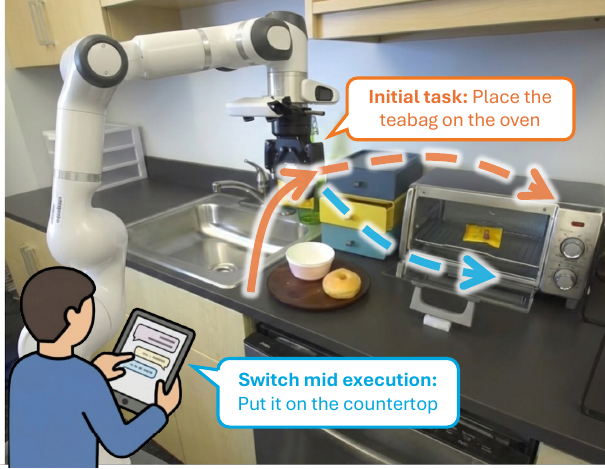}
    \vspace*{-7mm}
    \caption{Daily manipulation is diverse and time-critical, and often requires \textbf{interruptible} behavior, with users revising their intent after execution begins. This demands \textbf{interactive, steerable} policies that can switch behaviors \textbf{from any intermediate state} in response to a new task prompt. We propose \textit{ReSteer}, a framework to quantify and improve the steerability of multitask robot policies. }
    \label{fig:steerable_feature_img}
    \vspace*{-3mm}
\end{figure}

Steerability captures a policy's ability to properly condition on new task instructions in arbitrary states. 
Consider the example in \autoref{fig:steerable_feature_img}, where we initially instruct the robot to put the tea bag on the oven, then ask it to place it on the countertop. 
Existing policies~\citep{lee2025molmoact, intelligence2025pi05visionlanguageactionmodelopenworld, kim2024openvlaopensourcevisionlanguageactionmodel} tend to ignore the new instruction and continue executing the original task, operating as if they were conditioned on state alone.
This deficit is particularly problematic as it prevents users from issuing new instructions during ongoing execution, severely restricting when and where a robot can be meaningfully controlled.

In this work, we argue that a policy's steerability is a function of its training data distribution, with most existing datasets exhibiting pathologies that encourage learned models to ignore task conditioning. 
Concretely, existing datasets are collected subject to the following recipe: we identify a set of tasks, then collect demonstrations on each task independently. 
This process yields data where language is only informative for a small fraction of total task execution, leading to policies that over-index on state information to dictate behavior, a form of shortcut learning~\citep{Geirhos_2020}.
As an initial contribution, we formalize this intuition via conditional mutual information (CMI): given a dataset and learned policy, steerability can be approximated by $I(A; L | S)$. In other words, incorporating language prompts reduces the variance of the predicted action distribution, indicating that instructions meaningfully constrain action selection.
Empirically, we validate our proxy metric through experiments across multiple state-of-the-art VLAs, showing a strong correlation between our proposed metric and ``steerable" success rate.

Motivated by this diagnosis, we argue that incorporating data where the task instruction changes mid-execution---eliciting corresponding behavioral switches---is critical for improving the steerability of the learned policy. 
To this end, we introduce \textit{ReSteer}, a simple yet effective data generation framework that comprises two stages: \emph{stage-aware steering data generation} and \emph{self-refinement}.
In stage-aware steering data generation, for each switch from task A to task B, we
(i) use CMI to identify low-steerability states along task A rollouts,
(ii) pick a target state from task B, and
(iii) synthesize a transition trajectory between the two for training.
While this first stage introduces new task-switching trajectories, it does not always guarantee reliable execution, and the steering success rate can remain low.
To address this gap, we subsequently apply a self-refining behavioral cloning stage, where the policy is iteratively deployed to collect rollouts and finetuned solely on its own successful trajectories, thereby reinforcing in-distribution steerability induced by the newly injected steering data.
By combining these two stages, \emph{ReSteer} provides an efficient way to expand the steerability coverage and make steering behavior robust.

We evaluate \emph{ReSteer} through comprehensive experiments in both simulation and the real world.
In simulation on the LIBERO benchmark~\cite{liu2023liberobenchmarkingknowledgetransfer}, we observe that baseline policies exhibit limited steerability, with task switching feasible only early in execution and substantial performance degradation as the rollout progresses.
In contrast, \emph{ReSteer} achieves a \textbf{10\%} absolute improvement in successfully steering among the 10 benchmark tasks from diverse intermediate states.
These gains indicate that \emph{ReSteer} improves language steerability and makes mid-execution task switching more reliable in practice.
To isolate the contribution of each component of \emph{ReSteer}, we perform controlled ablations to analyze how sampling strategy and dataset size affect performance.
Furthermore, we conduct extensive real-world experiments in a daily kitchen scene, where \emph{ReSteer} improves steerability by $\mathbf{2.2\times}$ compared to a policy fine-tuned only on teleoperation data.
Lastly, we show that \emph{ReSteer}-enhanced policies can be chained to execute long-horizon task sequences. For example, a user can instruct the robot to switch from closing the oven to put the tea bag in the drawer, and then switch to pick up a bowl afterward (see \autoref{fig:DROID_quantitative}). This flexibility enables more interactive and reactive robotic assistants in real-world deployments.
Across our experiments, \emph{ReSteer} improves steerability, enabling more reactive multitask robot policies that better support human interaction.
We argue that quantifying and understanding steerability is important for reliable policy deployment, and \emph{ReSteer} offers a simple, practical way to do so.
We will release all code, data, and models to reproduce our results.

%% file: sections/02_related-work.tex
A rich body of work studies learning multitask, language-conditioned policies for robot control~\cite{Tellex_Kollar_Dickerson_Walter_Banerjee_Teller_Roy_2011, unknown}. While early approaches typically focus on following a fixed, task-specific instruction set, recent VLA policies are pretrained at scale and can generalize zero-shot to a much broader range of natural-language commands.

\noindent\textbf{From Language Conditioning to Interactive Prompting.}
Multitask policies such as BC-Z and RT-1/RT-X \citep{jang2022bcz,brohan2023rt1,openx} demonstrate that large-scale language-conditioned policies can solve many manipulation tasks from diverse instructions.
More recent works have targeted language following more directly.
Interleave-VLA~\citep{fan2025interleavevla} re-labels Open-X-Embodiment~\citep{openx} with interleaved image--text instructions to improve grounding of referring expressions.
ChatVLA/ChatVLA-2~\citep{zhou2025chatvla,zhou2025chatvla2} jointly train a VLM head and an action head to retain open-world reasoning while producing instruction-conditioned actions.
In interactive systems (e.g., Hi-Robot~\citep{hirobot}, Yell-at-your-robot~\citep{yellatrobot}, Robix~\citep{robix}), a VLM decomposes high-level user requests into short prompts for a downstream VLA.
SwitchVLA~\citep{switchvla} studies switching by conditioning on execution context, but relies on segmented phases and engineered behavior labels.
CAST~\citep{glossop2025castcounterfactuallabelsimprove} synthesizes counterfactual language for similar observations, improving navigation instruction following by increasing semantic diversity.
Overall, these methods either presume robust primitive instruction following or restrict switching to predefined modes; in contrast, we \emph{measure} and \emph{improve} language steerability under mid-episode prompt changes via targeted data generation and self-refinement.


\noindent\textbf{Steerability and Reasoning Beyond Language.} 
Most existing multitask approaches categorize control through two primary lenses: \textit{latent-space manipulation} and \textit{reasoning-based planning}. 
Latent-space methods utilize pretrained language-driven embeddings or internal latents~\citep{karamcheti2023voltron, li2025textlatent} to interpolate between behaviors. 
While effective for representation transfer~\citep{blade2024, NEURIPS2020_9909794d,wang2024discoveringroboticinteractionmodes}, these are often limited to predefined skill boundaries or treated as static inputs. 
Parallel work~\citep{zawalski2024ecot, duan2025fastecot, lee2025molmoact, gu2023rttrajectory} in reasoning-based planning augments policies with textual plans or spatial traces to improve execution. 
However, both directions generally assume a single fixed intent per rollout. 
Because they treat language as a static episode-level input or restrict task updates to predefined phase boundaries, they lack a mechanism for high-frequency, user-driven redirection from arbitrary intermediate states. 
\emph{ReSteer} addresses these drawbacks by treating steerability as an instant, state-dependent interruption, enabling a policy to remain interactive and responsive to intent changes at any point during execution.

\noindent\textbf{Our Perspective: Steerability of Multitask Policies.}
Later work on reasoning, interactive prompting, counterfactual relabeling, task switching, and text-latent control improves robustness and usability, but often assumes a single intent per episode, limits updates to predefined phase boundaries, or depends on external planners and engineered state machines.
In contrast, we target \emph{steerability}: a policy should be interruptible and re-steerable by language at any execution state, without fixed task boundaries or hand-designed switching modes.
Accordingly, we treat language following as a state-dependent property of the closed-loop policy and introduce evaluation protocols and training mechanisms designed specifically for this continuous notion of steerability.

%% file: sections/03_quantifying-steerability.tex





\subsection{Definitions}
\label{sec:quantifying:def}
Consider an MDP with state space $\StateSpace$, action space $\ActionSpace$,
and transition dynamics $\TransitionKernel$.
We assume a set of tasks $k \in \TaskIndexSet$, with corresponding language instruction $\LangInstr{k}$.
Let $\policy(a \mid s, \ell)$ denote a language-conditioned policy that outputs
an action distribution given state $s$ and instruction $\ell$.
For a fixed instruction $\LangInstr{k}$ and initial state distribution $p(s_0)$,
the closed-loop execution of $\policy$ induces a trajectory distribution
$
    \TrajDistPi{\policy}{k}
    \;\defeq\;
    p_\policy\bigl(s_0, a_0, s_1, a_1, \dots, s_T \mid \LangInstr{k}\bigr),
    \quad
    \traj = (s_0, a_0, \dots, s_T),
$
where $a_t \sim \policy(\cdot \mid s_t, \LangInstr{k})$.

We focus on pretrained policies that already exhibit strong multitask
performance under fixed instructions.
Let $\mathbb{I}_k(\traj)\in\{0,1\}$ denote the task-$k$ success indicator, and
define the \emph{success probability from state} $s$ as
\begin{equation}
    P^\policy_k(s)
    \;\defeq\;
    \Pr_{\traj \sim p_\policy(\cdot \mid s_0=s,\;\LangInstr{k})}
    \bigl[\mathbb{I}_k(\traj)=1\bigr]
\end{equation}

We assume an imitation learning setting, where expert demonstrations are collected under
instruction $\LangInstr{k}$, yielding trajectories
$\{\traj^{(i)}\}_{i=1}^{\NumDemos{k}}$ with $\NumDemos{k}$ demos.
We construct the imitation
dataset as the collection of state-action pairs along the recorded trajectories:
$
    \DemoSet{k}
    \;\defeq\;
    \Bigl\{(s_t^{(i)}, a_t^{(i)})\Bigr\}_{i=1,\dots,\NumDemos{k};\;t=0,\dots,T_i-1}
$. 
Let $\mu_k$ denote the empirical distribution over states visited in the expert
demonstrations of task $k$.
We assume the existence of thresholds $\alpha \in (0,1)$ and
$\delta \in [0,1)$ such that
\begin{equation}
    \Pr_{s \sim \mu_k}
    \bigl[P^\policy_k(s) \ge \alpha\bigr]
    \;\ge\;
    1-\delta,
    \qquad \forall k \in \TaskIndexSet
\end{equation}
This assumption ensures that the policy is competent at executing tasks from
in-distribution states, and that failures observed under mid-execution
instruction changes are attributable to limited steerability rather than poor
single-task execution.

\medskip
\noindent \textbf{State Visitation Sets.}
Given a policy $\policy$ and task $k$, we define the
\emph{policy-induced state visitation set} as the set of states from which the
policy can successfully complete task $k$ with high probability:
\begin{equation}
    \StateVisitSet{k}(\policy)
    \;\defeq\;
    \Bigl\{
        s \in \StateSpace
        \;\Big|\;
        P_k^{\policy}(s) \ge \alpha
    \Bigr\},
\end{equation}
where $P_k^{\policy}(s)$ denotes the task-$k$ success probability of policy
$\policy$ when initialized from state $s$, and $\alpha \in (0,1)$ is the success
threshold defined previously.

\medskip
\noindent \textbf{Union of Task Visitation Sets.}
For a task pair $i,j \in \TaskIndexSet$ and policy $\policy$, we define the
\emph{union of task visitation sets} as
\begin{equation}
    \StateUnionSet{i}{j}(\policy)
    \;\defeq\;
    \StateVisitSet{i}(\policy)
    \cup
    \StateVisitSet{j}(\policy).
\end{equation}
A state $s$ belongs to $\StateUnionSet{i}{j}(\policy)$ if the policy can
successfully complete \emph{at least one} of the two tasks when initialized from
$s$ under the corresponding instruction.
The union $\StateUnionSet{i}{j}(\policy)$ characterizes overall task feasibility,
while the intersection $\StateInterSet{i}{j}(\policy)$ identifies states where
both tasks are individually achievable and where steering between tasks is
well-defined.

\medskip
\noindent \textbf{Steerable States.}
For a task pair $(i,j)$ and policy $\policy$, we first define \emph{directional steerability}.
A state $s \in \StateSpace$ is said to be \emph{steerable from task $i$ to task $j$}
if, when the policy is executing task $i$ and the language instruction is switched
to $\LangInstr{j}$ at state $s$, the policy can still complete task $j$ with high
probability.
Formally, define the post-switch success probability
\begin{equation}
    P^{\policy}_{\rightarrow j}(s)
    \;\defeq\;
    \Pr_{\traj \sim p_\policy(\cdot \mid s_0=s,\;\LangInstr{j})}
    \bigl[\mathbb{I}_j(\traj)=1\bigr].
\end{equation}
We call $s$ steerable to $j$ if $P^{\policy}_{\rightarrow j}(s) \ge \alpha$.
The set of directionally steerable states from $i$ to $j$ is
\begin{equation}
    \mathcal{S}^{\mathrm{steer}}_{i \rightarrow j}(\policy)
    \;\defeq\;
    \Bigl\{
        s \in \StateSpace_i
        \;\big|\;
        P^{\policy}_{\rightarrow j}(s) \ge \alpha
    \Bigr\}.
\end{equation}

\noindent We further define \emph{bidirectional steerability} as:
\begin{equation}
    \mathcal{S}^{\mathrm{steer}}_{i \leftrightarrow j}(\policy)
    \;\defeq\;
    \mathcal{S}^{\mathrm{steer}}_{i \rightarrow j}(\policy)
    \;\cap\;
    \mathcal{S}^{\mathrm{steer}}_{j \rightarrow i}(\policy).
\end{equation}

\medskip
\noindent \textbf{Steerability Coverage Ratio (SCR).}
While bidirectional steerability characterizes whether instruction switching is possible at a given state, we seek a policy-level metric that quantifies how broadly such steering behavior is supported across the policy’s feasible state space.

We define the \emph{steerability coverage ratio (SCR)} for a task pair
$(i,j)$ and policy $\policy$ as
\begin{equation}
    \mathrm{SCR}_{i \leftrightarrow j}(\policy)
    \;\defeq\;
    \frac{
        \left|
            \mathcal{S}^{\mathrm{steer}}_{i \leftrightarrow j}(\policy)
        \right|
    }{
        \left|
            \StateUnionSet{i}{j}(\policy)
        \right|
    }.
\end{equation}

The denominator represents the full set of states at which the policy can
successfully execute at least one of the two tasks, and therefore remains fixed
under steerability-improving training. 
A higher SCR indicates that the policy can respond to task changes across a broader range of states during execution.

\medskip
\noindent \textbf{Data Dependence of Steerability.}
In standard multitask datasets, each trajectory is executed under a
single fixed instruction. Let $\mathcal{D}$ denote the training dataset, and let
$\policy_{\mathcal{D}}$ denote the policy obtained after training on
$\mathcal{D}$. Such data primarily determines the set of states from which the
trained policy can successfully execute at least one task. 
Formally, the dataset induces the \emph{union of task visitation sets} under the trained policy:
\begin{equation}
    \StateUnionSet{i}{j}(\policy_{\mathcal{D}})
    \;\defeq\;
    \StateVisitSet{i}(\policy_{\mathcal{D}})
    \;\cup\;
    \StateVisitSet{j}(\policy_{\mathcal{D}}),
\end{equation}

However, because $\mathcal{D}$ contains only single-instruction trajectories, it
provides no instruction-contrasting supervision at the same underlying state.
As a result, although $\StateUnionSet{i}{j}(\policy_{\mathcal{D}})$ may be large,
the bidirectionally steerable set
$\mathcal{S}^{\mathrm{steer}}_{i \leftrightarrow j}(\policy_{\mathcal{D}})$ can
remain small, leading to poor steerability coverage.

Improving steerability therefore requires augmenting $\mathcal{D}$ with
\emph{steering data} that explicitly injects paired supervision at shared
execution states. Concretely, for a state $s_t$, 
we add samples of the form $(s_t, a_t^{(i)}, \LangInstr{i})$ and $(s_t, a_t^{(j)}, \LangInstr{j})$,
which expose how actions should differ under the two instructions at the same
state. This instruction-contrasting augmentation enables the policy to learn
task-dependent switching behavior under mid-execution prompt changes, thereby
expanding $\mathcal{S}^{\mathrm{steer}}_{i \leftrightarrow j}(\policy_{\mathcal{D}})$.

\subsection{Evaluation of Steerability}
To evaluate steerability, we quantify the ability of a policy to complete any target task $\tau_j$ (specified by prompt $l_j$) from any trained states from task $\tau_i$.
A rollout is marked as successful if the policy, when conditioned on the provided new prompt $l_j$, completes the corresponding task $t_j$ according to the task’s success metric. 
To evaluate policy steerability at each task, we sample different timesteps to switch different task prompts during policy execution.
At each sampled state, we prompt the policy with all task prompts from the dataset.
In our experiment, we test each (state, prompt) pair with 10 rollouts ($n_{repeat}=10$) to account for policy stochasticity.
The whole evaluation requires $n\times(n-1)\times |k_i| \times n_{repeat}$ number of rollouts, where $k_i$ is the number of sampled states to steer from at a certain originally executing task $\tau_i$. Even though expensive, it provides a precise and exhaustive way to quantify steerability and directly matches the definition of steerability. We use this metric to compare against other proxies and benchmark different multitask policies.

\subsection{Case Study: Evaluating the Steerability of Multitask Policies}
\label{section:case study}
Using steerability evaluation, we analyze state-of-the-art VLA policies and reveal state-dependent failures to respond to instruction changes, motivating targeted data generation.
We quantify steerability for representative open-source VLA models:
$\pi_{0.5}$~\cite{intelligence2025pi05visionlanguageactionmodelopenworld}, OpenVLA-OFT~\cite{kim2025finetuningvisionlanguageactionmodelsoptimizing}, and MolmoACT~\cite{lee2025molmoact}. For each model, we evaluate steerability by switching
the language instruction mid-execution while keeping the environment state in-distribution, yielding a
controlled \emph{in-distribution steering} setting.
Because we focus on cross-task steering within a shared scene, we use LIBERO-Goal as the benchmark, a
commonly used small-scale multi-task dataset. Although all three VLAs achieve saturated single-task
performance on LIBERO-Goal (over $90\%$ success rate), none exhibits reliable cross-task steering under
prompt switches, as shown in \ref{table:baseline-libero}. This reveals a lack of \emph{in-domain} steerability in current VLAs, consistent with
policies that overfit to single-task behaviors despite high nominal success. 

\begin{table}[ht]
\centering
\small
\setlength{\tabcolsep}{6pt}
\begin{tabular}{l c c c}
\hline
 & \textbf{OpenVLA-OFT} & \textbf{MolmoACT} & $\boldsymbol{\pi_{0.5}}$ \\
\hline
\textbf{Steerability Score} & 0.252 & 0.295 & 0.403  \\
\hline
\end{tabular}
\caption{VLAs steerability comparison on LIBERO.}
\label{table:baseline-libero}
\vspace*{-4mm}
\end{table}

\begin{figure}[t]
    \centering
    \includegraphics[width=1.0\linewidth]{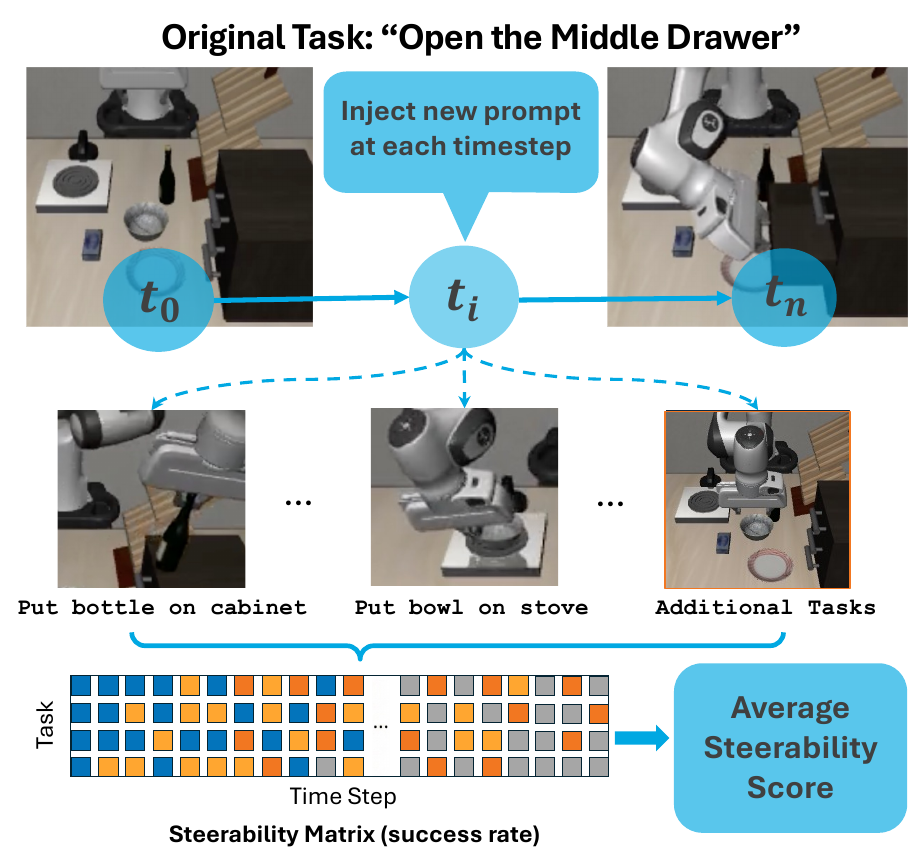}
    \vspace*{-7mm}
    \caption{Steerability evaluation. We run an original task $i$, then at sampled timesteps $t$ we switch the language prompt to a target task $j$ and measure whether the policy completes $j$ from that state. Repeating this over all sampled states and all target tasks yields a steerability matrix (time$\times$target task), whose average success rate gives the overall steerability score in task $i$.} 
    \label{fig:steerable_case_study}
    \vspace*{-3mm}
\end{figure}

\begin{figure*}[t]
    \centering
    \includegraphics[width=1.0\linewidth]{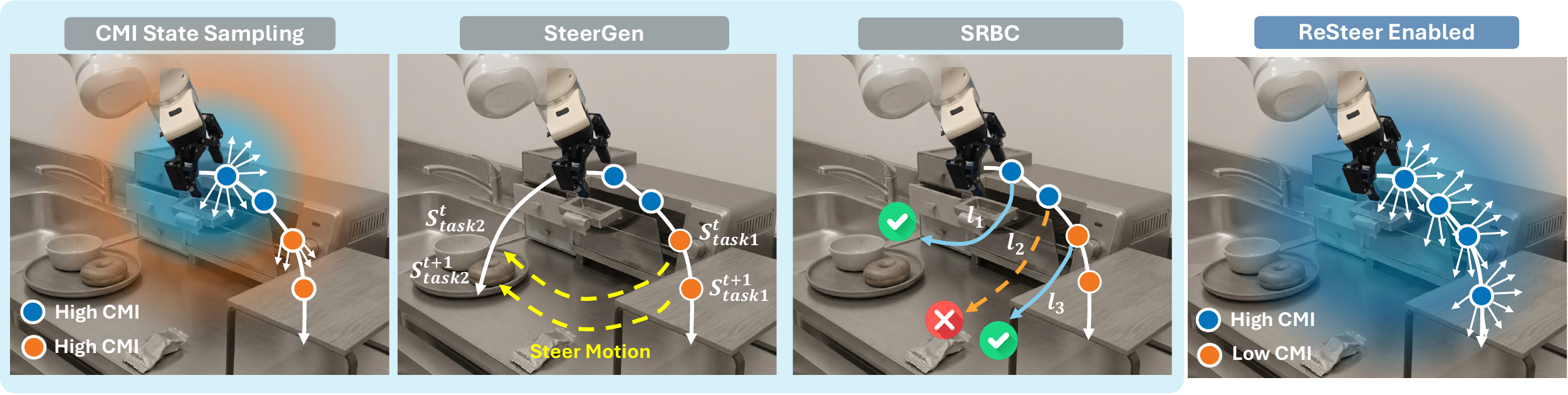}
    \vspace*{-7mm}
    \caption{We propose an online learning framework to improve the steerability of multitask policies. The framework comprises three components. First, we propose a CMI-based state sampling strategy that prioritizes data collection at the most unsteerable states, improving the sample efficiency of data generation. Second, we introduce a stage-aware steering data generation pipeline that synthesizes cross-task steering motions ($s^t_{task1} \rightarrow s^t_{task2}$), thereby expanding the steerable states, $\mathcal{S}^{\mathrm{steer}}_{i\leftrightarrow j}$. Third, we develop a self-refining behavior cloning (SRBC) scheme that finetunes the policy using successful steered trajectories given different instructions $l$ and consequently increases steerability coverage ratio $\mathrm{SCR}_{i \leftrightarrow j}(\policy)$.}
    \label{fig:steerable_method_img}
    \vspace*{-4mm}
\end{figure*}

%% file: sections/04_resteer-method.tex
\begin{figure}[t]
    \centering
    \includegraphics[width=1.0\linewidth]{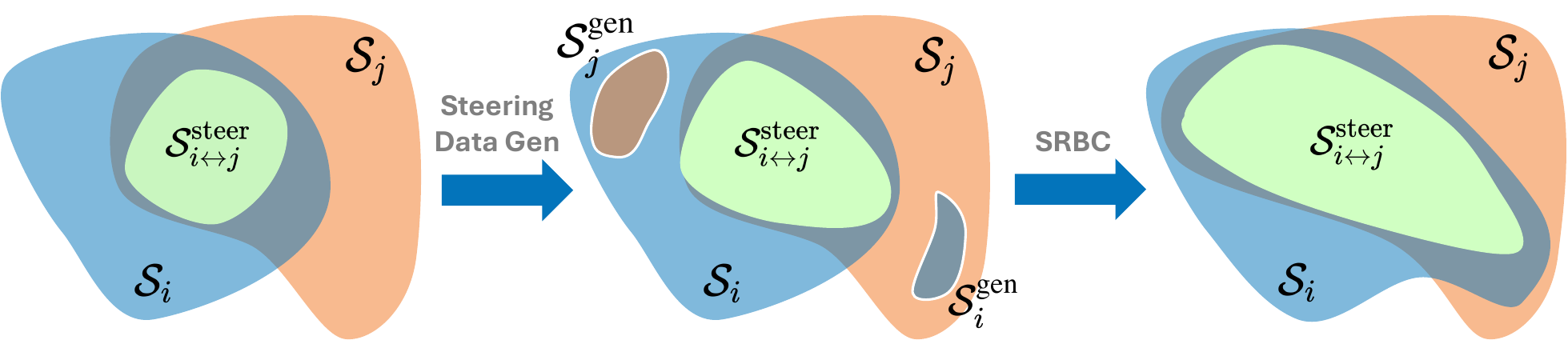}
    \vspace*{-5mm}
    \caption{Illustration of the steerability improvement afforded by ReSteer.
    Left: the bidirectionally steerable set $\mathcal{S}^{\mathrm{steer}}_{i\leftrightarrow j}$
    (green) occupies only a small subset of the policy-induced feasible states
    $\mathcal{S}_i$ and $\mathcal{S}_j$.
    Middle: stage-aware steering data generation ($\mathcal{S}^{\text{gen}}_i$ and $\mathcal{S}^{\text{gen}}_j$) introduces instruction-contrasting
    transitions, expanding the steerable region.
    Right: self-refining behavior cloning (SRBC) further enlarges steerability within
    the same feasible state space.
    }
    \label{fig:resteer_venn}
    \vspace*{-3mm}
\end{figure}

To improve the steerability coverage ratio (SCR), we show in the previous section that additional data is necessary. 
We therefore propose \emph{ReSteer}, a framework that progressively expands the steerable set through targeted data generation and policy refinement.

As illustrated in \autoref{fig:resteer_venn}, ReSteer first performs stage-aware steering data generation, which introduces instruction-contrasting transitions at compatible execution states. 
This expands the steerable sets $\mathcal{S}^{\mathrm{steer}}_{i\rightarrow j}(\policy)$ and increases the portion of feasible states where task switching is possible. 
However, exhaustively generating steering data for all task pairs and states is prohibitive, requiring $\mathcal{O}(n^2 k_i n_{\mathrm{r}})$ rollouts, where $n$ is the number of
tasks, $k_i$ denotes the number of states visited under task $i$, and
$n_{\mathrm{r}}$ is the number of rollout trials per state. 
To address this, \emph{ReSteer} leverages conditional mutual information (CMI) between action and language as a principled signal to identify states with low language–action coupling, where steering data is most needed.

Following this, \emph{ReSteer} applies self-refining behavior cloning (SRBC) to finetune the policy on its own successful steering rollouts. 
As shown in \autoref{fig:resteer_venn} (Right), this refinement enlarges the steerable region within the same feasible state space, further increasing SCR by strengthening in-distribution steering.

\subsection{SteerGen: Stage-Aware Steering Data Generation}
\label{sec:stage_aware_steering}

Prior work mitigates poor language steering through counterfactual data generation~\cite{glossop2025castcounterfactuallabelsimprove}, but has primarily been demonstrated in navigation settings and does not directly extend to long-horizon, contact-rich manipulation. 
Inspired by recent data-generation pipelines~\cite{mandlekar2023mimicgendatagenerationscalable,xue2025demogen}, we instead exploit task-stage structure to synthesize cross-task steering motions. 
Our \emph{stage-aware} procedure (i) segments demonstrations into semantic stages; (ii) samples a start state from a source task and selects a stage-matched end state from a target task by minimizing a shortest-path cost that favors short, smooth transitions; and (iii) generates a feasible
interpolation segment via motion planning.







\begin{figure*}[t]
    \centering
    \includegraphics[width=1.0\linewidth]{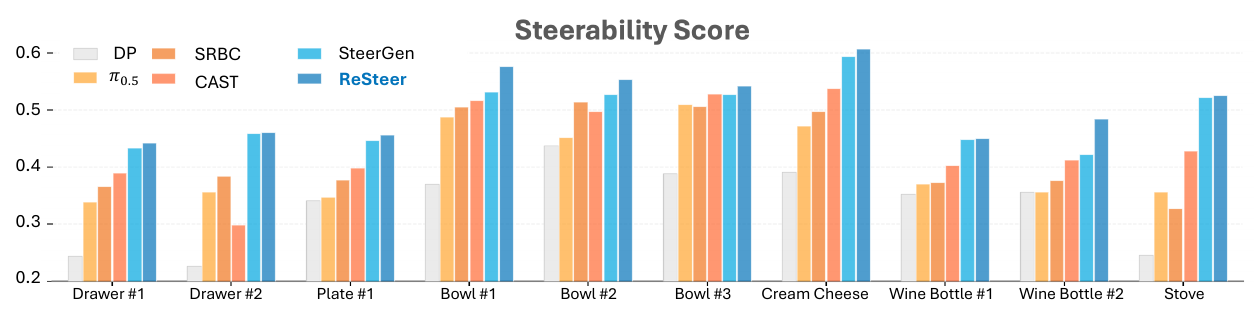}
    \vspace*{-7mm}
    \caption{\textbf{Evaluating Steerability on LIBERO-Goal.}
    For each source task (x-axis), we sample intermediate execution states and measure the average success rate of switching to each of the other nine target tasks under the corresponding language prompt (bars). 
    \emph{ReSteer} achieves the highest steerability across all ten tasks and consistently outperforms strong baselines such as CAST~\citep{glossop2025castcounterfactuallabelsimprove}, demonstrating the effectiveness of our two-stage data generation pipeline.
    }
    \vspace*{-3mm}
    \label{fig:steerable_sim_result_method_img}
\end{figure*}

\textbf{Stage-aware segmentation}
We assume a pick-and-place task structure and assign each state $s$ to one of three semantic stages:
\begin{equation}
    \sigma(s) \in \{\text{pre-grasp}, \text{transport}, \text{place}\}
\end{equation}
where $\sigma$ is a stage-labeling function.
In practice, $\sigma$ is computed using simple kinematic and contact cues,
including end-effector--object distance, gripper opening, and object support
height. The stage label $\sigma(s)$ is precomputed for each demonstration state.
These labels restrict steering to stage-consistent regions (e.g., pre-grasp
$\rightarrow$ pre-grasp across tasks), simplifying trajectory synthesis and
improving feasibility.

\textbf{Shortest-path end-state selection}
Steering data consists of instruction-contrasting supervision at the same
execution state, i.e., paired samples $(s,\LangInstr{i},a^{(i)})$ and
$(s,\LangInstr{j},a^{(j)})$ that specify how actions should change when the task
instruction is switched.
To obtain such data, we connect a state $s$ from a source-task trajectory to a
compatible state $s'$ from a target-task demonstration, after which execution
follows the original target rollout. 
This produces a feasible steering
trajectory that provides task-$j$ supervision at state $s$.

Given a start state $s$ and target task $\tau'$, we pick a stage-matched state
$s'$ with the same stage label
$\sigma(s')=\sigma(s)$ from target-task demonstrations such that the interpolation from $s$ to
$s'$ is shortest.


\textbf{Steering trajectory generation}
A steering trajectory consists of a short transition from the source state $s$ to the selected target state $s'$, followed by execution under the target instruction $\l_{\tau'}$. We instantiate this transition using simple linear interpolation.
On the physical system, the robot is reset to $s$, tracks the interpolation
segment using a low-level controller, and then executes the policy conditioned on $\l_{\tau'}$ from $s'$ onward. 
Successful executions are recorded as steering transitions and added to the training set for subsequent finetuning of $\policy$.

\subsection{Mutual-information-based Steering Trajectory Sampling}
As the number of tasks and execution horizon grow, exhaustively evaluating
steerability and generating steering trajectories at all states becomes
computationally infeasible. 
Moreover, steerability is highly non-uniform across the state space: only a subset of states consistently fail to respond to instruction changes. 
We therefore focus data collection on \emph{low-SCR} regions, prioritizing states where the policy is least responsive to language.

Since direct rollout-based evaluation at every state is impractical, we
introduce \emph{conditional mutual information} (CMI) between language and action
as an efficient offline proxy for steerability. 
At a fixed state $s$, the CMI is
defined as
\begin{equation}
    I(A; L \mid S = s)
    \;=\;
    H(A \mid S = s)
    \;-\;
    H(A \mid S = s, L)
    \label{eq:cmi}
\end{equation}
which quantifies the reduction in action uncertainty when the instruction is
known.

Intuitively, $H(A \mid S = s)$ captures the diversity of actions the policy may
produce from the same state across different instructions, while
$H(A \mid S = s, L)$ measures how deterministically the policy behaves under a
fixed instruction. 
A large CMI indicates that language strongly modulates the action distribution at state $s$, reflecting high steerability.
Conversely, low CMI implies weak language–action coupling, revealing states where instruction switches are unlikely to induce meaningful behavioral change.

\textbf{Theoretical Link: CMI as a Proxy for SCR}
To evaluate steerability without the computational burden of rollouts, we establish that CMI serves as a \textbf{necessary condition} for the Steerability Coverage Ratio (SCR). 
For a task pair $(i, j)$, the CMI at state $s$ is equivalent to the Jensen--Shannon (JS) divergence between instruction-conditioned action distributions:
\begin{equation}
    I^\pi_{i,j}(s) = \mathrm{JS}\left( \pi(\cdot \mid s, L_i) \,\|\, \pi(\cdot \mid s, L_j) \right)
\end{equation}
By applying \textit{Pinsker’s Inequality}, we can lower-bound this information by the squared Total Variation (TV) distance. Assuming a steerable state must exhibit a minimum behavioral shift $\varepsilon$ such that $\mathrm{TV}(\pi_i, \pi_j) \ge \varepsilon$, we obtain:
\begin{equation}
    s \in \mathcal{S}^{\mathrm{steer}}_{i \leftrightarrow j}(\pi) \implies I^\pi_{i,j}(s) \ge \tau, \quad \text{where } \tau = \tfrac{1}{2}\varepsilon^2
\end{equation}
This indicates that states with $I^\pi_{i,j}(s) < \tau$ are effectively ``instruction-blind'' and cannot support task switching. 
Let $\nu$ denote a reference distribution over $\StateUnionSet{i}{j}(\policy)$.
By taking the expectation over $s \sim \nu$, we establish that the SCR is upper-bounded by the probability of encountering high-CMI states:
\begin{equation}
    \mathrm{SCR}_{i \leftrightarrow j}(\pi) \le \Pr_{s \sim \nu} \left[ I^\pi_{i,j}(s) \ge \tau \right]
\end{equation}
This relationship justifies CMI as a \textit{rollout-free proxy} for identifying low-steerability regions without environment simulation. For the complete formal proof and further technical details, please refer to the Supplementary Material.

\textbf{CMI-Guided State Prioritization}
To operationalize the CMI defined in \autoref{eq:cmi}, we estimate the entropy terms using Gaussian kernel density estimation (KDE) \cite{guo2025demospeedupacceleratingvisuomotorpolicies}. For a state $s_t$, the conditional entropy $\widehat{H}(A \mid s_t, l)$ is computed from the empirical policy density $\widehat{p}(a \mid s_t, l)$:
\begin{equation}
\label{eq:cond_entropy}
    \widehat{H}(A_t \mid s_t, l) = -\frac{1}{NK} \sum_{i,j} \log \widehat{p}(a_{j}^{(i)} \mid s_t, l)
\end{equation}
where $a_{j}^{(i)}$ are action samples drawn from the policy. The marginalized entropy $\widehat{H}(A \mid s_t)$ is estimated by aggregating these samples across the set $\mathcal{L}$.

We use the resulting CMI to identify ``instruction-blind'' regions where the policy fails to differentiate behavior. Each state $s$ in the buffer $\mathcal{B}$ is assigned a weight $w(s) \propto g(\widehat{I}(A; L \mid s))$, where the shaping function $g$ is chosen to prioritize low-steerability (low-CMI) states. We then sample start states for new rollouts proportional to these weights. This mechanism allows us to trigger targeted steering rollouts \emph{online} during collection or curate high-difficulty states \emph{offline} before finetuning.

\subsection{Self-Refining Behavioral Cloning}
While stage-aware data generation increases task overlap by injecting new transitions, we observe that the resulting policy often fails to fully utilize these regions, leading to inconsistent task-switching. 
As illustrated in Fig.~\ref{fig:resteer_venn}, while the first stage expands the feasible intersection of tasks, the policy-induced steerable set $\mathcal{S}_{i\leftrightarrow j}^{\text{steer}}(\pi)$ may still only occupy a small portion of this region.

To close this gap and maximize the steerability coverage ratio (SCR), we propose \textit{Self-Refining Behavioral Cloning} (SRBC). Starting from the initial finetuned policy, we collect rollouts involving task switching and retain only the successful trajectories $\mathcal{D}_{\text{succ}}$. We then iteratively finetune the policy on the augmented dataset $\mathcal{D} \leftarrow \mathcal{D} \cup \mathcal{D}_{\text{succ}}$ by minimizing:
\begin{equation}
    \mathcal{L}_{\text{BC}}(\pi) = \mathbb{E}_{(o, a) \sim \mathcal{D}} [-\log \pi(a \mid o)]
\end{equation}
By reinforcing only successful steering behaviors, SRBC ``sharpens'' the policy's response to instruction changes, ensuring that the theoretical steerability provided by the generated data is reliably realized in practice.

\begin{figure}[h]
    \centering
    \includegraphics[width=1.0\linewidth]{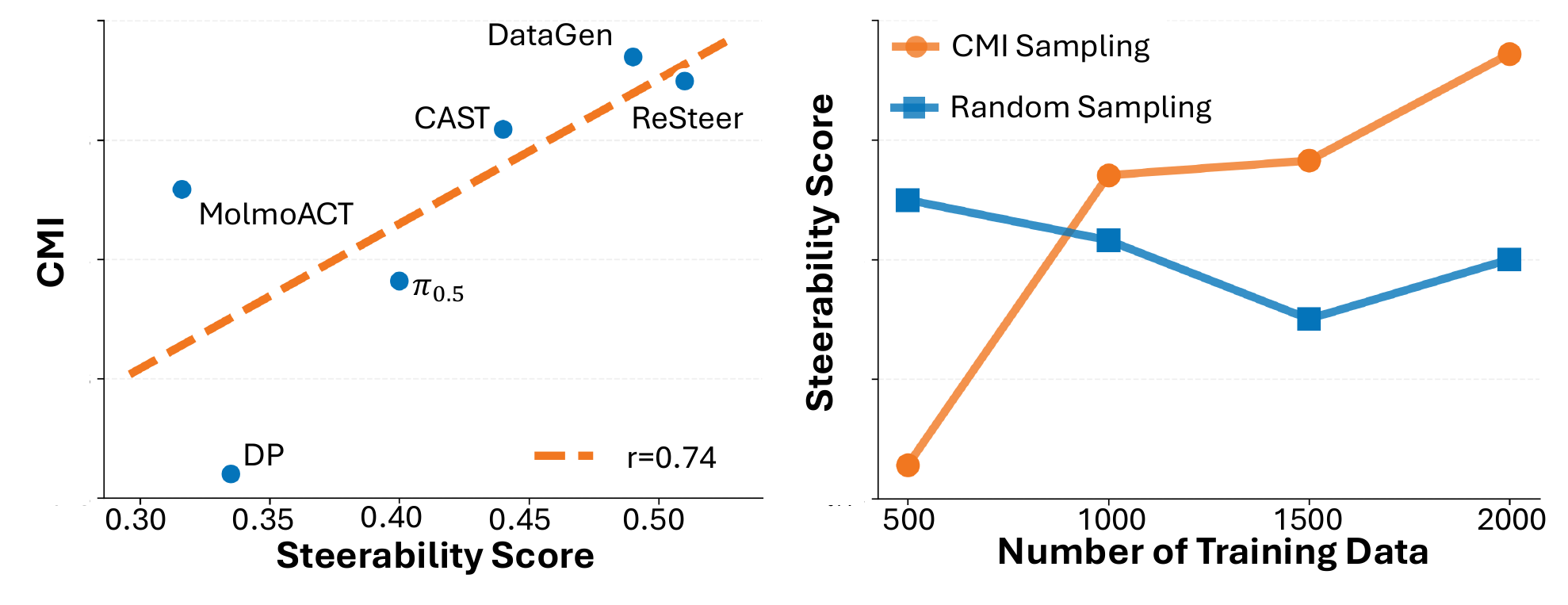}
    \vspace*{-7mm}
    \caption{To evaluate CMI as a proxy for steerability, we track CMI and steerability scores
across training checkpoints. As shown in left figure, CMI is positively
correlated with steerability, with particularly consistent trends within a
single model family (the $\pi_{0.5}$ variants). Motivated by this relationship, we use CMI to prioritize low-steerability states for data collection and
fine-tuning. Empirically, MI-guided sampling yields higher sample efficiency than uniform random sampling.}
    \label{fig:MI_result}
    \vspace*{-3mm}
\end{figure}

%% file: sections/05_sim-experiments.tex




\begin{figure}[h]
    \centering
    \includegraphics[width=1.0\linewidth]{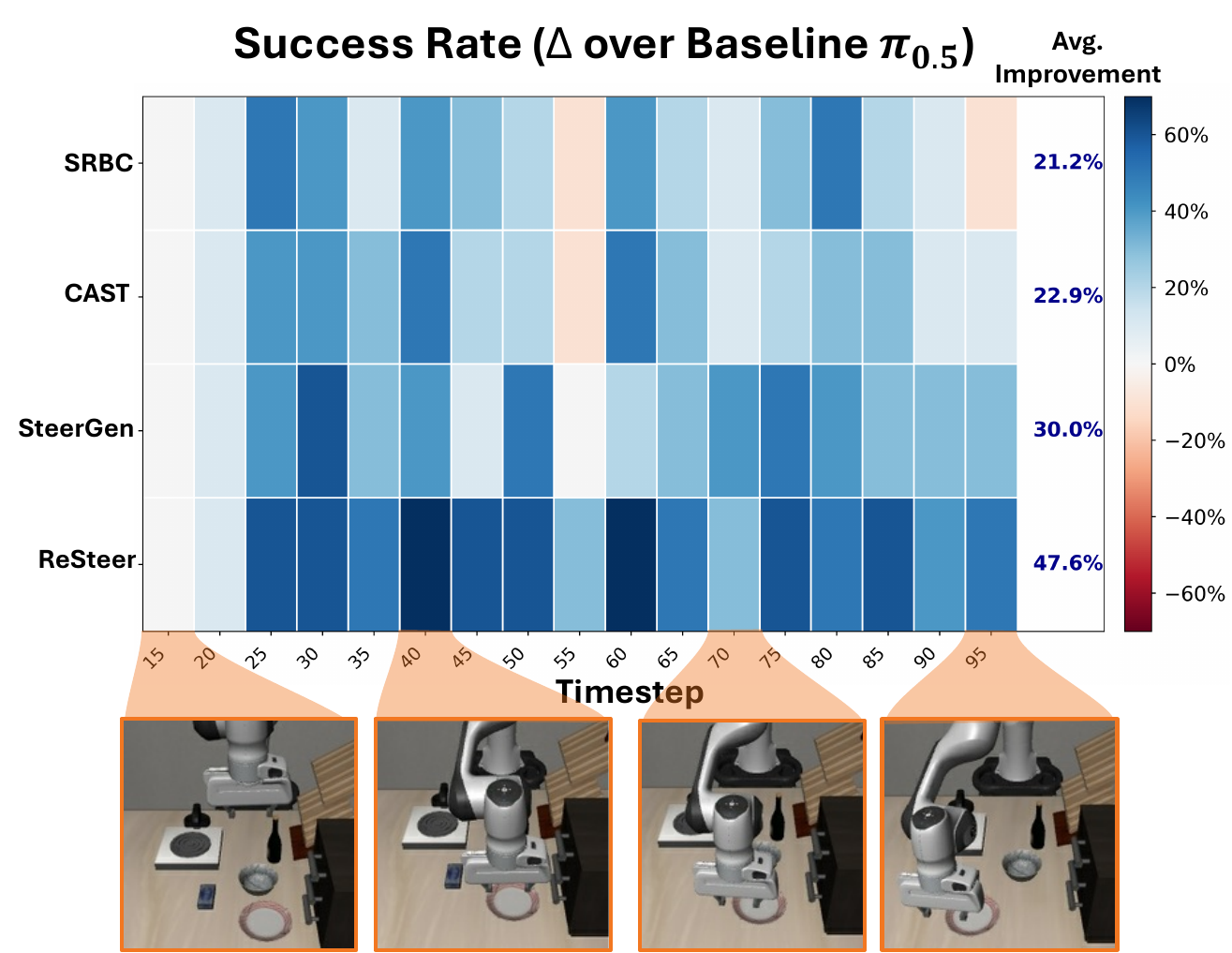}
    \vspace*{-5mm}
    \caption{We conduct a case study on task steering from \emph{Push the Plate to the Front of the Stove} $\rightarrow$ \emph{Put the Bowl on the Plate}. The prompt is switched at the indicated timestep, with representative LIBERO environment states shown at select timesteps. SteerGen and ReSteer consistently outperform all baseline methods, with particularly strong gains at later timestep switches. ReSteer further improves upon SteerGen by sharpening the policy in regions where SteerGen cannot consistently complete the task.}
    \label{fig:steerable_state}
\end{figure}

\begin{figure*}[t]
    \centering
    \includegraphics[width=1.0\linewidth]{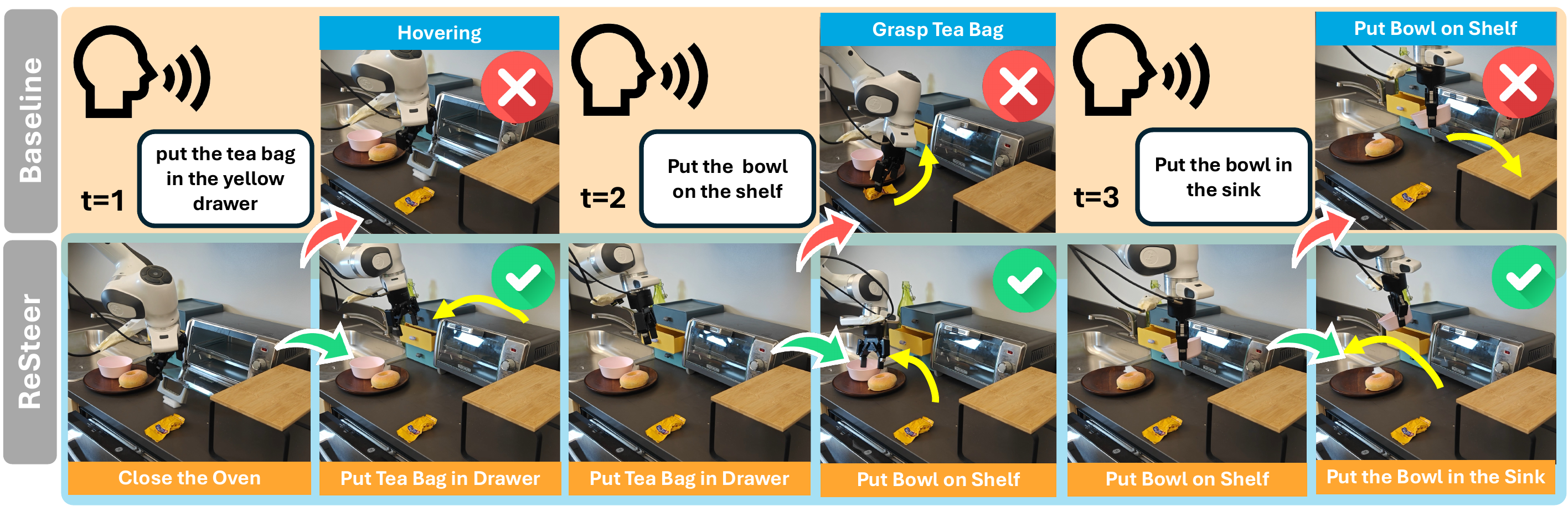}
    \vspace*{-7mm}
    \caption{In an example with four single tasks and three steering scenarios, ReSteer (bottom row) successfully switches to the requested next task upon user instruction, whereas the baseline (top row) exhibits failure modes such as getting stuck or continuing the original task (keep grasping the tea bag, put the bowl to the shelf instead of sink) despite the new prompt.}
    \label{fig:DROID_quantitative}
\end{figure*}

\begin{figure}[h]
    \centering
    \includegraphics[width=1.0\linewidth]{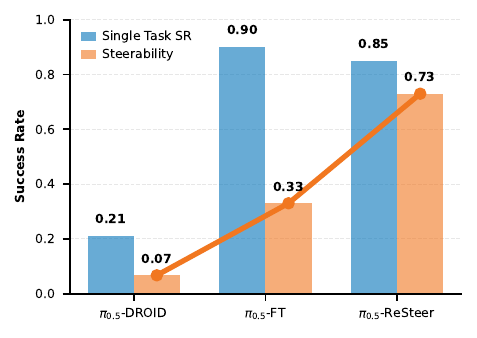}
    \vspace*{-9mm}
    \caption{We evaluate policies on the DROID platform across four single-task evaluations and three steering scenarios. In-domain teleoperation data improves single-task success over the baseline, but yields limited steerability. In contrast, ReSteer achieves $\mathbf{2.2\times}$ higher steerability while preserving single-task performance.}
    \label{fig:DROID_result}
    \vspace*{-3mm}
\end{figure}

We test the following hypotheses to illustrate the effectiveness of the proposed metrics and evaluation on steerability, as well as how our methods improve the steerability.
\begin{itemize}
    \item \textbf{H1}: Stage-aware data augmentation increases overlap between task state distributions, leading to improved in-domain steerability across tasks.
    \item \textbf{H2}: CMI serves as an effective proxy for steerability, reducing the need for exhaustive rollout-based evaluation.
    \item \textbf{H3}: CMI-guided state sampling improves the sample efficiency of steering data generation.
    \item \textbf{H4}: SRBC improves the success rate of steerable behaviors, bridging the gap between induced steerability from dataset overlap and realized steerability at execution time.
\end{itemize}

\subsection{Setup}
\textbf{Dataset}
We perform experiments with the LIBERO-Goal setup ~\cite{liu2023liberobenchmarkingknowledgetransfer}. LIBERO-Goal provides a single-scene environment containing versatile multi-task language labels, offering a controlled setting that is well-suited for evaluating steerability across tasks. We use $\pi_{0.5}$ as a base model and finetune it based on different settings.


\textbf{Baselines} 
We compare ReSteer against two representative multitask visuomotor learning approaches: 
(i) \textbf{Diffusion Policy (DP)} \cite{chi2023diffusionpolicy}, which we adapt for multitask control using FiLM-based instruction conditioning; and 
(ii) \textbf{CAST} \cite{glossop2025castcounterfactuallabelsimprove}, a data-augmentation baseline that improves instruction following by pairing existing observations with counterfactual language labels. 
To ensure a fair comparison, we implement a CAST-style variant using our stage-aware generator, extracting initial action snippets from generated steering motions to augment the training set with instruction-contrasting supervision.


\subsection{Results}
\textbf{H1: Stage-aware data augmentation improves steerability between tasks.}
Figure~\ref{fig:steerable_sim_result_method_img} shows that fine-tuning on our augmented data improves steerability by $8.8\%$, indicating that the resulting policy responds more reliably to language instructions across intermediate execution states than the baseline. We also observe that the CAST variant increases steerability; however, because it does not provide complete trajectories to realize full steering behaviors, it still underperforms ReSteer.


\textbf{H2: Conditional Mutual Information is a good proxy to reflect steerability.}
We evaluate the validity of CMI as a rollout-free proxy by examining its correlation with the ground-truth steerability score. 
As shown in Fig.~\ref{fig:MI_result}, we observe a strong positive correlation ($r=0.74$) across diverse model architectures and data augmentation strategies. This trend is even more pronounced within specific model families; for instance, the $\pi_{0.5}$ variants exhibit a Pearson correlation exceeding $0.9$. 
These results empirically validate that CMI effectively captures a policy's sensitivity to instruction changes, which matches our theory. 
This high correlation justifies using CMI as a principled signal to identify "instruction-blind" states and prioritize them for targeted data collection, significantly improving sample efficiency over uniform sampling.

\textbf{H3: MI-based sampling scheme improves the sample efficiency for data generation.}
Instead of augmenting states uniformly along each rollout, we concentrate data generation on low CMI states. 
We compare this CMI-guided strategy against random sampling, which selects task pairs $(a \!\rightarrow\! b)$ uniformly at random and uses the corresponding steering trajectories for training. 
We first collect $50$ short steering motion segments for each task pair by uniformly sampling from all task trajectories. 
Using a single rollout per task pair, we then estimate CMI for candidate states and construct two training sets of equal size: (i) a low-CMI subset obtained by down-sampling the segments with the smallest CMI, and (ii) a randomly down-sampled subset. Figure~\ref{fig:MI_result} (right) shows that CMI-guided sampling yields more consistent improvements in steerability as the dataset scales, indicating higher sample efficiency. 
In contrast, random sampling exhibits substantially higher variance.



\textbf{H4: SRBC sharpens the success rate of steerable tasks}.
We hypothesize that reinforcing the policy on its own \emph{successful} steering executions will increase robustness and thus improve overall steerability.
We evaluate \textit{Self-Refining Behavioral Cloning} (SRBC), which iteratively collects successful task-switching rollouts and finetunes the policy on these trajectories. 
In Figure~\ref{fig:steerable_sim_result_method_img}, we report steerability for (i) the base $\pi_{0.5}$ policy and (ii) our full pipeline, each with and without SRBC, keeping all other components fixed.
SRBC consistently improves steerability in both settings: it yields a robust gain for $\pi_{0.5}$ and a further gain when added to our method, achieving the highest overall success rate. 
The improvement is consistent across tasks, indicating that SRBC provides a general, reliable boost to steerability rather than a task-specific effect.



%% file: sections/06_real-experiments.tex


We conduct real-world experiments using the DROID~\cite{khazatsky2025droidlargescaleinthewildrobot} platform and the $\pi_{0.5}$ DROID checkpoint. The scene is a kitchen-style setup designed to reflect everyday use. It contains an oven, a countertop shelf, a plate, and a bowl, enabling multiple manipulation tasks and within-episode task switching.

\subsection{Task Design and Data Collection}


\textbf{Teleoperated Demonstrations.}
We construct four different manipulation tasks based on the kitchen setup:
\begin{itemize}
    \item \textbf{Place bowl on shelf:} grasp the bowl, put it on the shelf.
    \item \textbf{Place bowl in sink:} grasp the bowl, put it in the sink.
    \item \textbf{Store tea bag in drawer:} open the yellow drawer, grasp the tea bag, and close the drawer.
    \item \textbf{Close oven:} grasp the oven door and close it.
\end{itemize}
We curate the finetuning datasets by teleoperating the robot and collecting a dataset of expert trajectories.

\textbf{\emph{ReStreer} Data Generation.}
Specifically, we sample states from the same motion stage across different demonstrations and generate steering motions between those states. Due to real-world efficiency constraints, we do not run SRBC in our hardware experiments.

To illustrate how \emph{ReStreer} enables more interactive manipulation with multitask policies, we design a long-horizon task sequences composed of the tasks described above, with mid-execution task steering:
\begin{enumerate}
    \item Close oven. After closing, steer to (from Place stage):
    \item Put the tea bag in the yellow drawer. After opening the drawer, steer to (from Pregrasp stage):
    \item Put the white bowl in the sink. After picking up the bowl, steer to (from Transport stage):
    \item Put the white bowl to the countertop shelf.
\end{enumerate}


\subsection{Results}

\vspace{4pt}
\begin{table}[t]
\centering
\small
\setlength{\tabcolsep}{4.5pt}
\renewcommand{\arraystretch}{1.15}
\begin{tabular}{l c c c}
\hline
\textbf{Test Case} & $\boldsymbol{\pi_{0.5}}$ & $\boldsymbol{\pi_{0.5}}$-FT & $\boldsymbol{\pi_{0.5}}$-\emph{ReSteer} \\
\hline
\multicolumn{4}{l}{\textbf{Single-task success (\%)}} \\
\hline
t1-close the oven & 0 & \textbf{100} & 80 \\
t2-put tea bag in drawer & 4 & 60 & \textbf{70} \\
t3-put white bowl on the shelf & 20 & \textbf{100} & \textbf{100} \\
t4-put white bowl in the sink & 60 & \textbf{100} & \textbf{100} \\
\hline
\multicolumn{4}{l}{\textbf{Steering success (\%)}} \\
\hline
\textbf{Pregrasp} t1$\rightarrow$t3 & \textbf{100} & 0 & 80 \\
\textbf{Pregrasp} t3$\rightarrow$t1 & 0 & 40 & \textbf{100} \\
\textbf{Pregrasp} t2$\rightarrow$t3 & 0 & 20 & \textbf{100} \\
\textbf{Transport} t3$\rightarrow$t4 & 20 & \textbf{80} & \textbf{80} \\
\textbf{Place} t1$\rightarrow$t2 & 0 & \textbf{60} & \textbf{60} \\
\hline
\end{tabular}
\caption{Success rates for single-task execution and pregrasping-stage steering across three methods.}
\label{table:single-and-steering-success}
\vspace*{-4mm}
\end{table}

We evaluate \emph{ReSteer} on the DROID platform across four manipulation tasks to determine if it can resolve the steerability failures observed in state-of-the-art policies. 
As illustrated in Fig.~\ref{fig:DROID_result}, while standard teleoperation finetuning ($\pi_{0.5}$-FT) saturates single-task success at 90\%, it yields only marginal improvements in steerability (33\%) compared to the 7\% of the zero-shot baseline. 
This indicates a ``specialization pathology'' where the model over-indexes on state information to complete a task, effectively ignoring mid-execution instruction changes. 
In contrast, \emph{ReSteer} achieves a 73\% steering success rate---a $2.2\times$ improvement over the finetuned baseline---while maintaining high single-task proficiency (85\%). 
Table~\ref{table:single-and-steering-success} reveals that the most significant improvements occur during \textit{pregrasp-stage transitions}, such as $t3 \rightarrow t1$ (white bowl on shelf $\rightarrow$ close oven) and $t2 \rightarrow t3$ (tea bag in drawer $\rightarrow$ white bowl on shelf), where \emph{ReSteer} achieves 100\% success compared to the baseline's 40\% and 20\%. 
These peak gains suggest that \emph{ReSteer} successfully addresses high-conflict decision points where the robot must break its commitment to an initial object to re-orient toward a new goal.


%% file: sections/07_discussion.tex

% ...

\textbf{Limitations.} Even though we provide a detailed and thorough analysis on steerability of VLA in a same test scene setting and illustrate how data distribution affects the steerability of the model. Such analysis is costly to run in real-world and test under different scenarios. 


\textbf{Future Works.} A promising direction is \emph{online interactive learning}: during deployment, the robot can accept natural-language corrections and use MI-style signals to decide when to query the user, log corrective segments, and prioritize updates on low-steerability states. To keep this practical and safe, we will study lightweight on-robot adaptation.

%% file: sections-appendix/x_overview.tex

\section*{Supplementary Material}


The supplementary material is structured as follows:
\begin{itemize}
    \item \textbf{ReSteer Framework and Algorithm} (\autoref{sec:app:system_overview}): this section describes the detailed workflow of the \emph{ReSteer} framework and implementation of each component.
    \item \textbf{Steerability Plot and Comparison for Tasks} (\autoref{sec:app:steerability_and_viz}): this section provides steerability plots for individual tasks and further analysis.
    \item \textbf{Analysis on Conditional Mutual Information Result} (\autoref{sec:app:CMI_analysis})
    \item \textbf{Visualization and Discussion of LIBERO Trajectories}: we visualize the LIBERO tasks' trajectories to discuss the potential loss of steerability when finetuned on a smaller dataset.(\autoref{sec:app:vis_libero})
    \item \textbf{Inspecting Language Information Restoration from VLA Latent} (\autoref{sec:app:inspect_latent})
    \item \textbf{Properties of Steerable States} (\autoref{sec:app:properties}): we derive the properties of steerable states based on the definitions.
    \item \textbf{Proof: CMI is a Proxy for Steerability} (\autoref{sec:app:cmi_proxy}): this section shows a proof sketch and theoretical link between CMI and steerability.
    \item \textbf{Hyperparameters} (\autoref{sec:app:hyperparam})
\end{itemize}


\section{System Overview}
\label{sec:app:system_overview}

\textbf{ReSteer Framework.}
ReSteer is an iterative data-centric refinement procedure that
progressively expands and consolidates the steerable state set of
a pretrained multitask policy. The framework alternates between
two stages. First, it performs CMI-guided steering data generation
(\emph{SteerGen}), where states with weak language–action coupling
are identified and augmented with instruction-contrasting transitions
to enlarge the feasible steering region. Second, it applies
self-refining behavioral cloning (SRBC), which reinforces successful
task-switching rollouts collected from the updated policy to improve
the reliability of steering behavior. By repeatedly injecting
targeted steering supervision and reinforcing realized switching
success, ReSteer increases steerability coverage while preserving
single-task competence.

We detail the \emph{ReSteer} framework in \autoref{algo:framework}.
\begin{algorithm}[h!]
\caption{ReSteer Framework}
\begin{algorithmic}[1]
\Require Pretrained policy $\pi$, Dataset $\mathcal{D}$, Number of Action Samples $N_a$
\While{True}

    $\mathcal{D}_{\text{SteerGen}=\{\}}$
    
    Sample states from task $i$: $s_0, s_1, \ldots, s_t \;\sim\; p^{\pi}_k(\tau \mid s_0)$
    \For{$s_0, s_1, \ldots, s_t$} \Comment{Calculate CMI for rollout States}
        \State Sample $N_a$ action chunks for each $l_j$: $a_j \sim p(a|s_t,l_j)$
        \State Calculate $\widehat{I}(A;L \mid s_t)$ \Comment{\autoref{eq:cond_entropy}} 
    \EndFor
    \For{$s_i,l_j\sim q(s_t,l)$} \Comment{\autoref{eq:sampling_weight}: Sampling from Low CMI States}
        \State Select end state $s_{j}$
        \State $\tau_j^i=\text{SteerGen}(s_i, s_j)$
        \State $\mathcal{D}_{\text{SteerGen}} \gets \tau_j^i $
    \EndFor
    \State $\pi_{\text{SteerGen}} \gets$ Finetune($\pi$, $\mathcal{D} \cup \mathcal{D}_{\text{SteerGen}}$)
    \State Sample $n$ rollouts: $\tau_{success} \sim \pi_{\text{ReSteer}}$
    \State $\mathcal{D}_{\text{SRBC}} \gets \tau_{success}$
    \State $\pi_{\text{ReSteer}} \gets$ Finetune($\pi$, $\mathcal{D} \cup \mathcal{D}_{\text{SteerGen}} \cup \mathcal{D}_{\text{SRBC}}$)
    \State $\pi = \pi_{\text{ReSteer}}$
\EndWhile
\end{algorithmic}
\label{algo:framework}
\end{algorithm}

\textbf{CMI-Guided State Prioritization.}
In Algorithm~\ref{algo:framework} (Lines 3–7), we estimate
the conditional mutual information (CMI) at each rollout state
to identify instruction-blind regions where steering supervision
is most needed.

To compute the conditional action entropy
$H(A \mid S=s_t,\,L=l)$, we approximate the instruction-conditioned
action density using Gaussian kernel density estimation (KDE)~\cite{guo2025demospeedupacceleratingvisuomotorpolicies}.
Given $N_a$ sampled action chunks of horizon $K$ under instruction $l$,
the density at timestep $t$ is estimated as
\begin{equation}
    \widehat{p}(A_t \mid s_t, l)
    =
    \frac{1}{N_a K h}
    \sum_{j=t}^{t+K-1}
    \sum_{i=1}^{N_a}
    \frac{1}{\sqrt{2\pi}}
    \exp\!\left(
        -\frac{\|a_j - a_{j}^{(i)}\|_2^2}{2 h^2}
    \right),
    \label{eq:kde-density}
\end{equation}
where $h$ is the KDE bandwidth.

The conditional entropy at state $s_t$ is then estimated as
\begin{equation}
    \widehat{H}(A_t \mid s_t, l)
    =
    -\frac{1}{N_a K}
    \sum_{j=t}^{t+K-1}
    \sum_{i=1}^{N_a}
    \widehat{p}\!\left(a_{j}^{(i)} \mid s_t, l\right)
    \log
    \widehat{p}\!\left(a_{j}^{(i)} \mid s_t, l\right).
    \label{eq:cond-entropy}
\end{equation}

To estimate the language-marginalized entropy $H(A_t \mid s_t)$,
we aggregate samples across prompts $l' \in \mathcal{L}$
using the same KDE procedure.
The empirical CMI used in Line~4 of the algorithm is then
\[
\widehat{I}(A;L \mid s_t)
=
\widehat{H}(A_t \mid s_t)
-
\widehat{H}(A_t \mid s_t, L).
\]

In practice, instead of flattening the entire action chunk
(which leads to high-dimensional density estimation),
we compute entropy using the difference between the final action
in the chunk and the current robot state.
Empirically, this lower-dimensional representation yields
more stable density estimates.

\textbf{Sampling States Based on CMI.}
After computing $\widehat{I}(A;L \mid s_t)$ for rollout states
(Line~4), we assign each candidate pair $(s_t,l)$
a sampling weight,
\begin{equation}
    w(s_t,l) \;\propto\; g\!\left(\widehat{I}(A;L \mid s_t)\right),
\end{equation}
where $g(\cdot)$ is a shaping function that emphasizes
low-steerability states (low CMI).

As shown in Line~6 of Algorithm~\ref{algo:framework},
start states for steering generation are sampled from
\begin{equation}
\label{eq:sampling_weight}
    q(s_t, l)
    =
    \frac{w(s_t,l)}
    {\sum_{s' \in \mathcal{B}}
     \sum_{l \in \mathcal{L}} w(s',l)},
\end{equation}
where $\mathcal{B}$ is a buffer of states collected across tasks.

This CMI-guided prioritization focuses steering data generation
on states where the policy exhibits weak language–action coupling,
thereby improving sample efficiency and accelerating expansion
of the steerable set.


\section{Steerability Plot and Comparison for Tasks}
\label{sec:app:steerability_and_viz}

\begin{figure*}[h]
    \centering
    \includegraphics[width=1.0\linewidth]{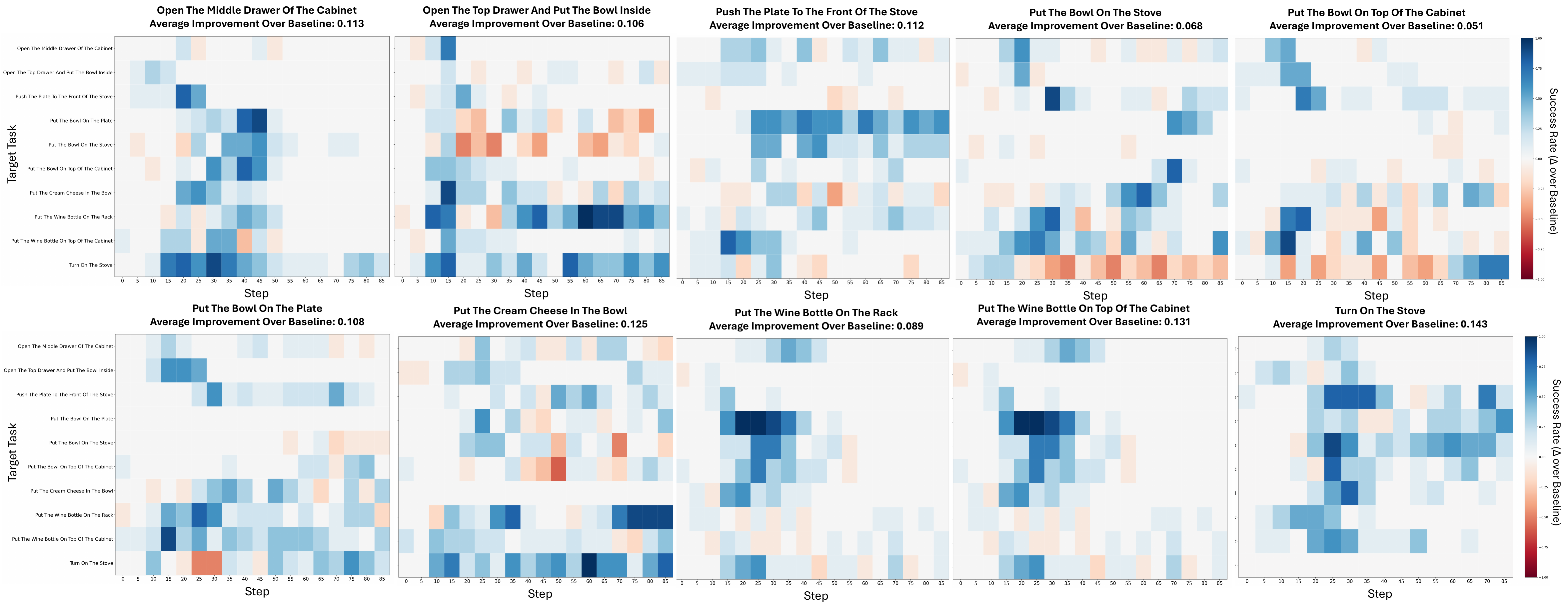}
    \vspace*{-7mm}
    \caption{
Task-level steerability improvement on LIBERO-Goal.
Each subplot corresponds to a source task (executed initially),
and reports the improvement in steering success rate
($\Delta$ over the baseline policy) when switching to all other
target tasks at different timesteps.
The x-axis shows the execution timestep of task switching,
the y-axis lists target tasks,
and color encodes the success-rate difference, where blue indicates improvement.
The number above each subplot indicates the average improvement
across all target tasks for that source task
ReSteer consistently improves steerability across tasks,
with particularly large gains at later timesteps where task commitment
is stronger and baseline policies tend to ignore prompt changes.
}
    \label{fig:steer_success}
\end{figure*}

Figure~\ref{fig:steer_success} visualizes steerability
improvement across all LIBERO-Goal tasks.
For each source task, we execute the original instruction and
sample intermediate states along the rollout.
At each sampled timestep, we switch the prompt to every other
target task and measure the steering success rate.
The x-axis denotes the timestep at which the task switch occurs during execution,
the y-axis enumerates the target tasks,
and the color of each cell indicates the difference in success rate, with blue regions corresponding to improvement in success rate.


\paragraph{Temporal Structure of Steerability.}
Across tasks, we observe a consistent temporal pattern:
steerability improvement is modest at early timesteps and
becomes more pronounced at later timesteps.
This trend reflects the fact that early execution states
often remain compatible with multiple task objectives,
whereas later states exhibit stronger task commitment
(e.g., object grasping or contact),
making mid-execution switching more challenging.
ReSteer yields particularly large gains in these high-conflict regions,
demonstrating that the generated steering data effectively expands
the feasible switching set.

\paragraph{Task-Dependent Asymmetry.}
The magnitude of improvement varies across source tasks.
Tasks involving object placement or contact-rich manipulation
(e.g., \emph{Turn On The Stove}, \emph{Put The Wine Bottle On Top Of The Cabinet})
show the largest average gains (up to $0.143$).
These tasks require decisive behavioral redirection,
and baseline policies tend to over-commit to the original goal.
ReSteer substantially mitigates this commitment bias.

In contrast, tasks with less constrained object geometry
(e.g., \emph{Put The Bowl On Top Of The Cabinet})
show smaller improvements, suggesting that these tasks
already exhibit moderate baseline steerability.

\paragraph{Implications for Steerability Coverage.}
The consistent positive $\Delta$ across all source tasks
indicates that ReSteer enlarges the steerable state set
without sacrificing feasibility.
Importantly, gains are not confined to specific task pairs;
improvements are broadly distributed across targets,
suggesting that the framework enhances general
language–action coupling rather than overfitting to
particular transitions.

Overall, the visualization confirms that ReSteer
systematically improves mid-execution prompt switching,
with the largest benefits emerging in states where
baseline policies exhibit strong task commitment.

\begin{figure*}[h]
    \centering
    \includegraphics[width=1.0\linewidth]{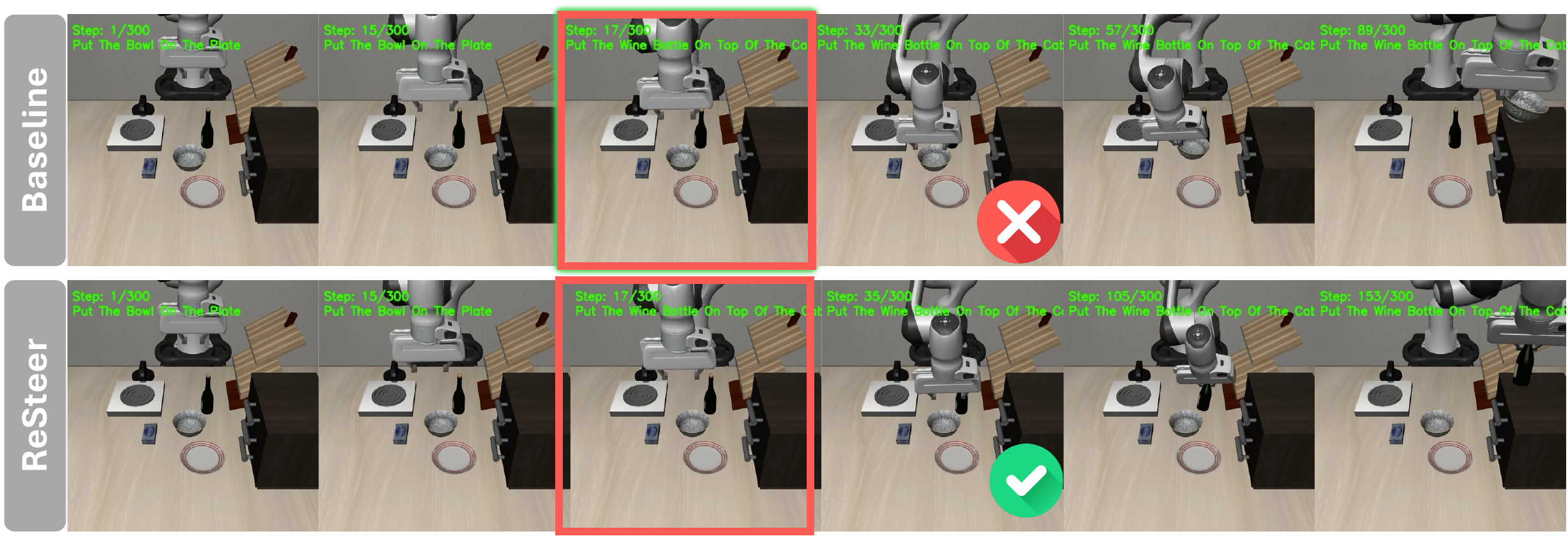}
    \vspace*{-7mm}
    \caption{We prompt the model to switch from the "put the bowl on the stove" to "put the wine bottle on top of the cabinet" at $t=17$. While baseline follows "put the bowl on the stove" throughout the rollout, \emph{ReSteer} manages to switch to the execution of the "put the wine bottle on top of the cabinet."}
    \label{fig:LIBERO_quantitative}
\end{figure*}


\section{Analysis on Conditional Mutual Information Result}
\label{sec:app:CMI_analysis}




\begin{figure*}[h]
    \centering
    \includegraphics[width=1.0\linewidth]{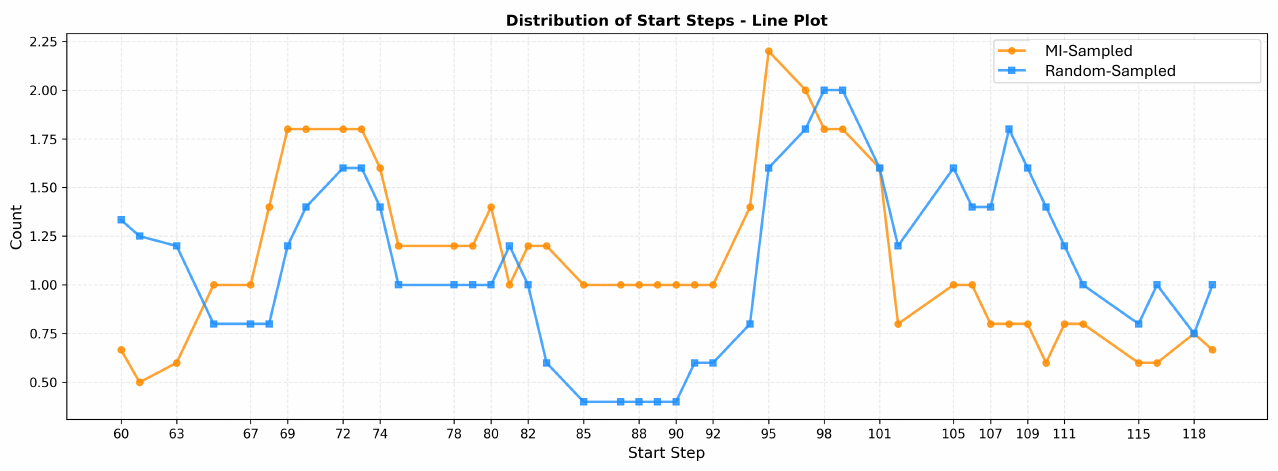}
    \vspace*{-7mm}
    \caption{
Comparison of start-step distributions under MI-guided sampling
and uniform random sampling.
The x-axis denotes the timestep selected as the starting state
for steering data generation, and the y-axis shows the sampling count.
MI-guided sampling concentrates selections in specific temporal
regions corresponding to low-CMI (instruction-blind) states,
where language–action coupling is weak.
In contrast, random sampling distributes selections more uniformly
across the rollout.
This targeted concentration demonstrates the improved sample
efficiency of CMI-based prioritization, focusing data generation
on states that most require steering supervision.
}
    \label{fig:DROID_CMI_quantitative}
\end{figure*}

\begin{figure*}[h]
    \centering
    \includegraphics[width=1.0\linewidth]{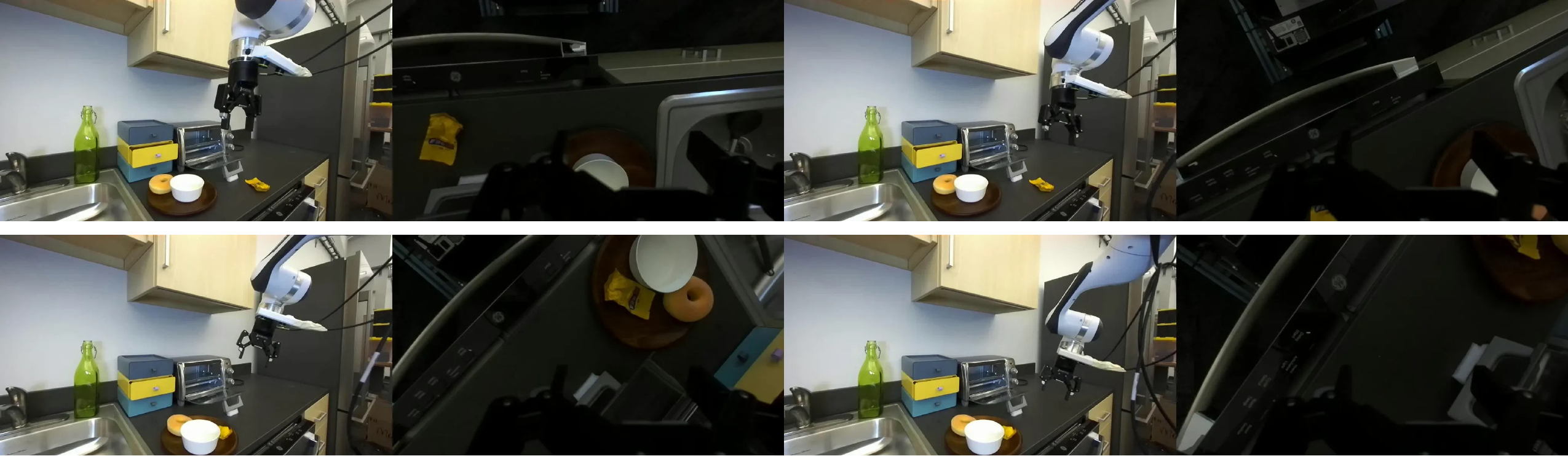}
    \vspace*{-7mm}
    \caption{
Visualization of MI-guided sampled states on the DROID platform.
Highlighted frames correspond to rollout states selected using the
CMI-based prioritization strategy.
The sampled states concentrate around high-conflict or
decision-critical phases (e.g., object approach, grasp transition,
and pre-placement), where baseline policies exhibit weak
language–action coupling.
This confirms that CMI-guided sampling identifies instruction-blind
regions in real-world trajectories and focuses data generation
on states that most require steering supervision,
thereby improving sample efficiency compared to uniform sampling.
}
    \label{fig:DROID_CMI_qualitative}
\end{figure*}

\begin{figure}[h]
    \centering
    \includegraphics[width=0.75\linewidth]{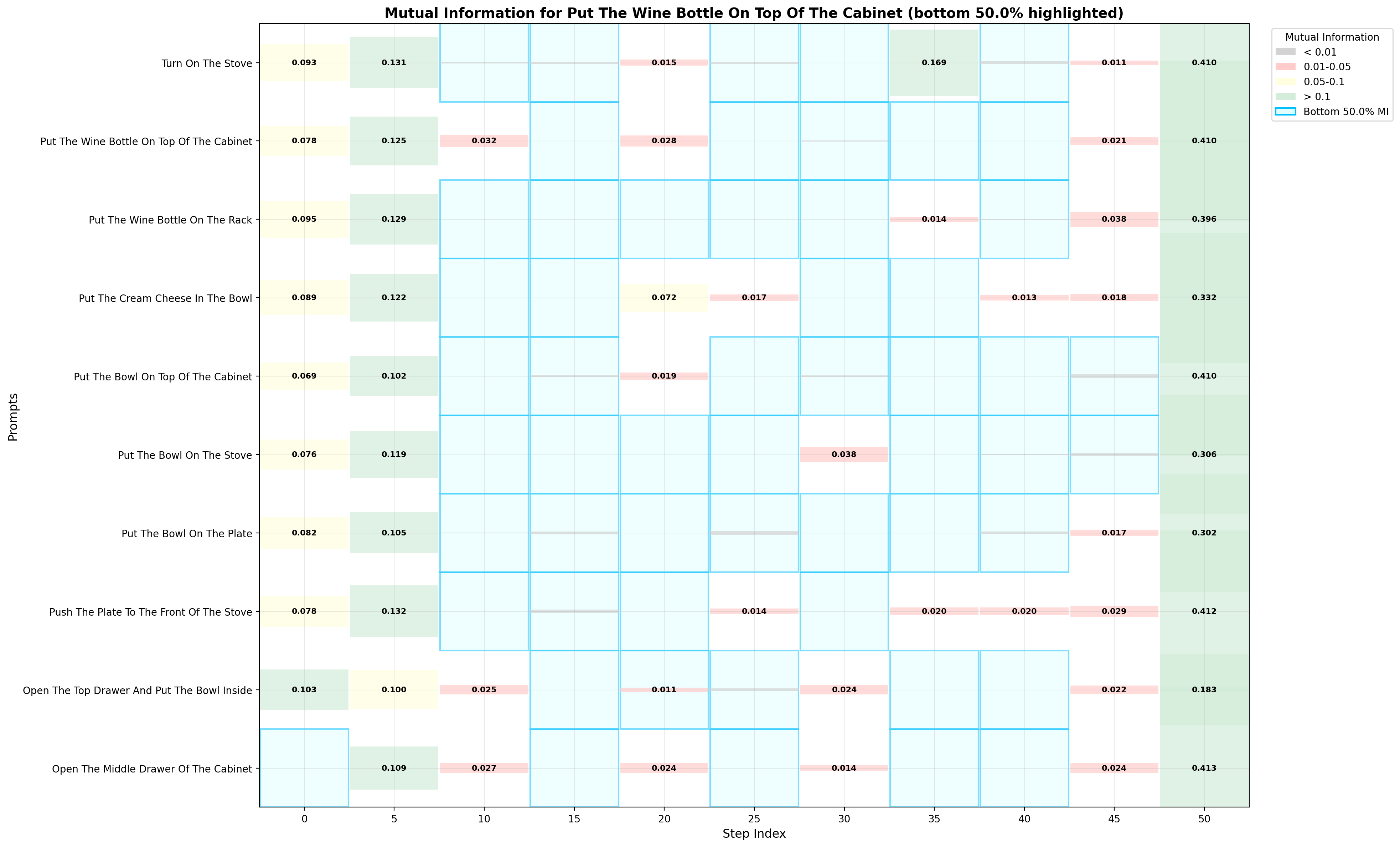}
    \caption{Visualization of states that were sampled according to the CMI Metric for rollout of task \"put the wine bottle on top of the cabinet". The blue boxes indicate the states selected with mutual information.}
    \label{fig:MI_state_libero}
\end{figure}


Conditional mutual information (CMI) serves as a necessary
indicator of language-sensitive behavior: a policy that ignores
language will satisfy $I(A;L\mid S_t)\approx 0$.
However, our empirical results reveal that large conditional MI
alone does not guarantee successful or task-aligned steering.

Specifically, we observe two failure modes in which CMI can
be misleading:

\paragraph{High influence but low competence.}
A model may exhibit large marginal action entropy $H(A\mid S_t)$
and correspondingly large $I(A;L\mid S_t)$ because different
prompts shift the action distribution toward distinct but
incorrect modes. In this regime, the policy is \emph{sensitive}
to language, yet the induced behaviors are not aligned with
task success. The mutual information is inflated by
unstructured or noisy variability rather than meaningful
goal-directed control.

\paragraph{Confident collapse to incorrect actions.}
Conversely, conditioning on language may sharply reduce
$H(A\mid S_t, L)$, producing a deterministic but incorrect
action. In this case, $I(A;L\mid S_t)$ can remain nontrivial
even though the policy is consistently wrong. Here,
language influences behavior, but not in a task-consistent
manner.

To better quantify the fraction of action uncertainty resolved
by language, we introduce a normalized variant:
\begin{equation}
\label{eq:norm-cmi}
\widetilde{I}(A;L \mid S_t)
\;=\;
\frac{I(A;L \mid S_t)}{H(A\mid S_t)}.
\end{equation}
This normalization discounts states with inherently large
action variance and more directly reflects the relative
strength of language conditioning.

\paragraph{Visualization of MI-Based State Selection.}
~\autoref{fig:MI_state_libero} visualizes the CMI values
along a rollout of the task \emph{“put the wine bottle on top of the cabinet.”}
Each row corresponds to a target prompt, and each column
corresponds to a timestep.
Cells are color-coded by MI magnitude, and blue boxes
highlight the bottom 50\% of states selected for steering
data generation.

Several structural patterns are evident.
First, low-MI states cluster in contiguous temporal segments,
particularly during mid-rollout phases where the policy’s
behavior is dominated by state dynamics (e.g., reaching or
transport) rather than instruction-dependent branching.
These states are precisely where baseline policies exhibit
instruction-blind behavior.

Second, high-MI regions tend to occur near decision points
(e.g., object interaction or goal placement), where language
meaningfully modulates the action distribution.
However, as discussed above, not all high-MI states correspond
to successful steering; some reflect unstable or misaligned
behavioral divergence.

The CMI-guided sampling mechanism in
Algorithm~\ref{algo:framework} therefore focuses on the
instruction-blind regions (blue-highlighted states),
injecting instruction-contrasting supervision exactly where
the baseline policy fails to differentiate tasks.
This targeted expansion of language–action coupling improves
steerability coverage while avoiding redundant data generation
in already language-sensitive regions.

\paragraph{CMI-Guided Sampling for DROID Experiment}
Based on the above observation, we compare CMI-guided state prioritization against uniform random sampling on the DROID platform.
We use SteerGen to collect 150 steering trajectories for switching from \emph{close the oven} to \emph{put the white bowl in the sink}.
We then construct two equal-sized training sets by downsampling to 50 trajectories: (i) a CMI-guided subset that prioritizes low-CMI (instruction-blind) start states, and (ii) a uniformly random subset.
\autoref{fig:DROID_CMI_quantitative} reports the resulting start-timestep distributions and we visualize representative start states selected by CMI sampling \autoref{fig:DROID_CMI_qualitative}.
Finally, we fine-tune separate policies on the two subsets and evaluate steering from \emph{close the oven} to \emph{put the white bowl in the sink}.
CMI-guided sampling achieves higher steering success (80\%) than random sampling (75\%), suggesting better sample efficiency by focusing supervision on states where language--action coupling is weakest.

\begin{table}[h]
\centering
\small
\setlength{\tabcolsep}{6pt}
\begin{tabular}{l c c}
\hline
 & \textbf{CMI-Sampling} & \textbf{Random Sampling} \\
\hline
\textbf{Steerability Score} & 0.8 & 0.75 \\
\hline
\end{tabular}
\caption{Steerability Score for Different Sampling Strategy on DROID}
\label{table:DROID-sampling}
\vspace*{-4mm}
\end{table}


\section{Visualization and Discussion of LIBERO Trajectories}
\label{sec:app:vis_libero}
\begin{figure}[h]
    \centering
    \includegraphics[width=1.0\linewidth]{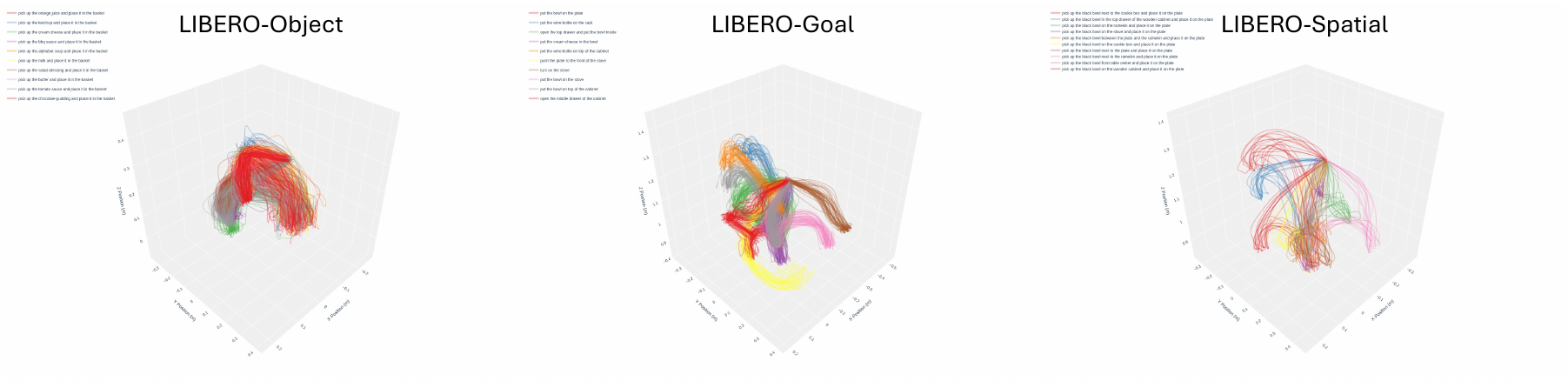}
    \vspace*{-5mm}
    \caption{We visualize the LIBERO benchmark trajectories. Most of the task trajectories from the dataset are clustered into a small space and lead to a relatively small $\mathcal{S}^\text{steer}$, which potentially makes the learn policies less steerable.}
    \label{fig:libero_trajs}
    \vspace*{-3mm}
\end{figure}

Visualization of LIBERO Trajectories.
Figure~\autoref{fig:libero_trajs} visualizes the end-effector trajectories for LIBERO tasks. We observe that tasks exhibiting substantial spatial overlap in end-effector trajectories tend to be more amenable to steering. Intuitively, high overlap suggests shared underlying motion primitives, which facilitates transferring control signals across tasks.

Discussion.
Tasks that are already closely related in the original dataset (e.g., put bowl on plate and put bowl on stove) naturally exhibit high steerability. As a result, the relative gains from our method on these tasks are smaller, since much of the transferable structure is already present. In contrast, tasks with less initial overlap benefit more from steering, leading to larger relative improvements.

\section{Inspecting Task Information Restoration from VLA Latent}
\label{sec:app:inspect_latent}
We study whether semantic information associated with task instructions is already encoded in the visual representation learned by the policy. Concretely, we test whether different language prompts are \emph{linearly separable} from visual latents alone, without access to language embeddings. A positive result would indicate that the policy can infer task identity directly from visual context, providing evidence for a shortcut in which language conditioning is partially ignored.

\textbf{Hypothesis.}
Visual observations (and their latent representations) may contain sufficient semantic cues to identify the task being executed. As a result, the policy can recover task intent from vision alone, reducing reliance on explicit language input.

\subsection{Experiment 1: Multitask Linear Separability from Visual Latents}

\begin{figure}[h]
    \centering
    \includegraphics[width=0.75\linewidth]{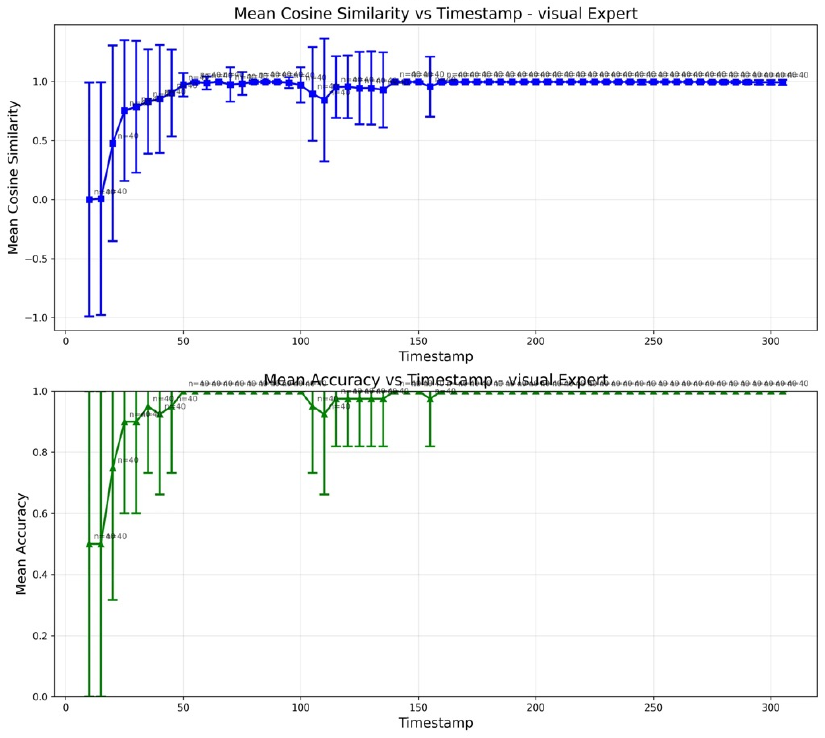}
    \vspace*{-5mm}
    \caption{\textbf{Stage-dependent linear separability of task semantics from visual latents.}
    We report the performance of a linear probe trained to predict task identity from visual latents as a function of rollout timestep.
    \textbf{Top:} mean cosine similarity between predicted and ground-truth task prototypes.
    \textbf{Bottom:} mean classification accuracy.
    Early in the rollout (pre-grasp stage), both metrics are low and exhibit high variance, indicating that visual latents are not linearly separable across tasks when the scene is visually similar.
    As execution progresses and physical interaction occurs (post-grasp and transport stages), both cosine similarity and accuracy rapidly increase and saturate near $1.0$, with reduced variance.
    This demonstrates that task semantics become strongly encoded in visual latents once task-specific effects are reflected in the environment state, enabling task inference from vision alone.}
    \label{fig:LIBERO_OBJ_latent}
    \vspace*{-3mm}
\end{figure}

\textbf{Setup.}
We consider a multitask setting with $K=10$ tasks in LIBERO-Object. For each task, we collect visual latents produced by the policy encoder during closed-loop rollouts. Each latent corresponds to a visual observation $o_t$ and is extracted \emph{without} conditioning on language embeddings.

To stress-test linear separability, we aggregate latents across multiple sources of variation:
(i) different task identities,
(ii) different rollout timesteps,
and (iii) multiple camera viewpoints (agent-view and wrist camera).
We explicitly exclude the right camera, which is not consistently available across all rollouts.

We train a linear classifier (logistic regression) to predict the task identity from the visual latent. This is the hardest probing setting, as the classifier must remain invariant to time, viewpoint, and trajectory-specific variations.

\textbf{Metrics.}
We report (i) classification accuracy on a held-out evaluation set and
(ii) cosine similarity between predicted and ground-truth class prototypes, which reflects the geometric separability of task clusters in latent space.

\textbf{Results.}
Training on agent-view and wrist-camera latents ($\sim$500 training samples, $\sim$100 evaluation samples), the linear probe achieves an accuracy of $95.8\%$ and an average cosine similarity of $0.91$.
This indicates that task identity is highly linearly decodable from visual latents alone, even under substantial nuisance variation.

\textbf{Interpretation.}
These results suggest that, across different scenes and execution stages, the visual latent implicitly encodes task semantics. Consequently, the policy can often infer ``what to do'' from vision alone, which helps explain why language conditioning becomes weak or redundant in multitask execution.

\subsection{Experiment 2: Single-Task Prompt Separability Across Execution Stages}

\textbf{Setup.}
We next analyze a more controlled setting to understand \emph{when} visual latents encode instruction-specific information. We focus on a single task executed in a fixed scene and collect visual latents from two different trajectories that share the same initial setup but diverge due to different language prompts.

We assign different labels to the two trajectories and train a linear probe to distinguish them based solely on visual latents. This isolates prompt-related effects while holding task identity and scene constant.

\textbf{Results.}
As shown in \autoref{fig:LIBERO_OBJ_latent}, we find that visual latents are \emph{not} linearly separable during early execution stages (e.g., pre-grasp), where observations are nearly identical across prompts.
However, after physical interaction occurs (e.g., post-grasp or object motion), linear separability emerges.

\textbf{Interpretation.}
This stage-dependent behavior indicates that visual latents encode prompt-specific semantics only after the environment state begins to reflect the consequences of the instruction. Prior to interaction, language-conditioned differences are not visually observable, forcing the policy to rely on language input. After interaction, the visual state itself reveals task intent, enabling prompt inference from vision alone.

\textbf{Summary.}
Together, these experiments show that visual latents can strongly encode task semantics, especially in multitask settings and later execution stages. This supports the hypothesis that VLA policies may over-index on visual context and under-utilize language, contributing to reduced steerability under mid-execution prompt changes.


\section{Properties of Steerable States}
\label{sec:app:properties}
The steerable states definitions in \autoref{sec:quantifying:def} imply several structural relationships between task competence, feasibility, and steerability.

\textbf{Steerability implies task competence.}
For any policy $\policy$ and task pair $(i,j)$,
\begin{equation}
    \mathcal{S}^{\mathrm{steer}}_{i \rightarrow j}(\policy)
    \;\subseteq\;
    \StateVisitSet{j}(\policy).
\end{equation}
\emph{Explanation.}
Directional steerability from $i$ to $j$ requires that, after switching the
instruction to $\LangInstr{j}$, the policy can complete task $j$ with high
probability. Therefore, any steerable state must also be a state from which task
$j$ is individually feasible.

\textbf{Bidirectional steerability implies joint competence.}
For any task pair $i,j \in \TaskIndexSet$ and a policy $\policy$, we define the
\emph{intersection of task visitation sets} as
\begin{equation}
    \StateInterSet{i}{j}(\policy)
    \;\defeq\;
    \StateVisitSet{i}(\policy)
    \cap
    \StateVisitSet{j}(\policy).
\end{equation}
A state $s$ belongs to $\StateInterSet{i}{j}(\policy)$ if the policy can
successfully complete \emph{both} tasks $i$ and $j$ with high probability when
initialized from $s$ under their respective language instructions.
For any policy $\policy$,
\begin{equation}
    \mathcal{S}^{\mathrm{steer}}_{i \leftrightarrow j}(\policy)
    \;\subseteq\;
    \StateInterSet{i}{j}(\policy)
    \;=\;
    \StateVisitSet{i}(\policy)\cap \StateVisitSet{j}(\policy).
\end{equation}
\emph{Explanation.}
Bidirectional steerability requires that the policy successfully execute both
tasks after instruction switches in either direction. Consequently,
steerability is only well-defined at states where both tasks are individually
achievable, namely the states contained in the intersection of task visitation
sets. This constraint fundamentally distinguishes steerability from
spatial or temporal generalization: rather than requiring robustness across
unseen states or time steps, steerability concerns the policy’s ability to
adapt its behavior when the task instruction is changed at a specific state
during execution.

\textbf{Steering is more restrictive than task feasibility.}
In general,
\begin{equation}
    \mathcal{S}^{\mathrm{steer}}_{i \rightarrow j}(\policy)
    \;\subseteq\;
    \StateVisitSet{j}(\policy),
    \qquad
    \mathcal{S}^{\mathrm{steer}}_{i \leftrightarrow j}(\policy)
    \;\subseteq\;
    \StateInterSet{i}{j}(\policy),
\end{equation}
with strict inclusion commonly observed in practice.
\emph{Explanation.}
Although a policy may be capable of executing a task from a given state, it may
fail to adapt its behavior when the instruction is changed mid-execution.
Empirically, we observe many states where both tasks are feasible, yet
instruction switching does not reliably alter the policy’s behavior, resulting
in non-steerable states within the feasible region.

\textbf{Steering is undefined outside the union of task visitation sets.}
For any policy $\policy$,
\begin{equation}
    \mathcal{S}^{\mathrm{steer}}_{i \rightarrow j}(\policy)
    \;\subseteq\;
    \StateUnionSet{i}{j}(\policy)
    \;\defeq\;
    \StateVisitSet{i}(\policy)\cup \StateVisitSet{j}(\policy).
\end{equation}
\emph{Explanation.}
Steering presupposes that at least one of the tasks is executable at the current
state. Outside the union of task visitation sets, neither task can be completed,
and instruction switching is therefore ill-defined.

\textbf{Directional asymmetry.}
Directional steerability is not symmetric in general:
\begin{equation}
    \mathcal{S}^{\mathrm{steer}}_{i \rightarrow j}(\policy)
    \;\neq\;
    \mathcal{S}^{\mathrm{steer}}_{j \rightarrow i}(\policy).
\end{equation}
\emph{Explanation.}
Switching from task $i$ to task $j$ may succeed while the reverse transition
fails, due to asymmetries in task structure, environment dynamics, or policy
representations. For example, transitioning from a pre-grasping behavior to a
placing behavior may be feasible, while reversing the instruction after placing
may not restore a valid grasping configuration.

\section{Proof: CMI is a proxy for steerability}
\label{sec:app:cmi_proxy}
We now formalize the relationship between state-conditional mutual information
and steerability coverage.

Without the loss of generality, fix a task pair $(i,j)$ and a policy $\policy$. For a given state $s$, recall that
the instruction-conditioned action distributions are given by
$\policy(\cdot \mid s, \LangInstr{i})$ and $\policy(\cdot \mid s, \LangInstr{j})$.
We define the state-conditional mutual information between actions and task
instructions as
\begin{equation}
    I^\policy_{i,j}(s)
    \;\defeq\;
    I(A;L \mid S=s),
\end{equation}
restricted to the two instructions $\LangInstr{i}$ and $\LangInstr{j}$.

\paragraph{Mutual information and action distribution separation.}
For a fixed state $s$, the conditional mutual information admits the closed-form
expression
\begin{equation}
    I^\policy_{i,j}(s)
    \;=\;
    \mathrm{JS}\!\left(
        \policy(\cdot \mid s, \LangInstr{i})
        \;\middle\|\;
        \policy(\cdot \mid s, \LangInstr{j})
    \right),
\end{equation}
where $\mathrm{JS}(\cdot\|\cdot)$ denotes the Jensen--Shannon divergence between
the two instruction-conditioned action distributions.

By Pinsker’s inequality, Jensen--Shannon divergence lower-bounds the squared
total variation distance. Specifically, for any two probability measures $p$ and
$q$,
\begin{equation}
    \mathrm{JS}(p\|q)
    \;\ge\;
    \tfrac{1}{2}\,
    \mathrm{TV}(p,q)^2 ,
\end{equation}
where $\mathrm{TV}(p,q)=\tfrac{1}{2}\|p-q\|_1$ denotes total variation distance.
Applying this inequality yields
\begin{equation}
    I^\policy_{i,j}(s)
    \;\ge\;
    \tfrac{1}{2}\,
    \mathrm{TV}\!\left(
        \policy(\cdot \mid s, \LangInstr{i}),
        \policy(\cdot \mid s, \LangInstr{j})
    \right)^2 .
\end{equation}

\paragraph{Steerability implies instruction-dependent action separation.}
We assume the existence of a constant $\varepsilon > 0$ such that for any state
$s$ that is bidirectionally steerable between tasks $i$ and $j$, the policy must
exhibit a nontrivial change in its action distribution under instruction
switching:
\begin{equation}
    s \in \mathcal{S}^{\mathrm{steer}}_{i \leftrightarrow j}(\policy)
    \;\Longrightarrow\;
    \mathrm{TV}\!\left(
        \policy(\cdot \mid s, \LangInstr{i}),
        \policy(\cdot \mid s, \LangInstr{j})
    \right)
    \;\ge\;
    \varepsilon .
\end{equation}
This assumption formalizes the requirement that successful task switching must
induce meaningfully different control behaviors at the switching state.

Combining the above inequalities, we obtain the implication
\begin{equation}
    s \in \mathcal{S}^{\mathrm{steer}}_{i \leftrightarrow j}(\policy)
    \;\Longrightarrow\;
    I^\policy_{i,j}(s)
    \;\ge\;
    \tau,
    \qquad
    \tau \defeq \tfrac{1}{2}\varepsilon^2 .
\end{equation}
Equivalently,
\begin{equation}
    I^\policy_{i,j}(s) < \tau
    \;\Longrightarrow\;
    s \notin \mathcal{S}^{\mathrm{steer}}_{i \leftrightarrow j}(\policy).
\end{equation}

\paragraph{Implication for steerability coverage.}
Let $\nu_{i,j}$ denote a reference distribution over feasible states (e.g.,
uniform over the task union set). Taking expectation over $s \sim \nu_{i,j}$
yields
\begin{equation}
    \mathrm{SCR}_{i \leftrightarrow j}(\policy)
    \;\le\;
    \Pr_{s \sim \nu_{i,j}}
    \!\left[
        I^\policy_{i,j}(s) \ge \tau
    \right].
\end{equation}

This result establishes that state-conditional mutual information provides a
necessary-condition proxy for steerability: states with low mutual information
cannot be reliably steered between tasks, and the fraction of high-mutual-
information states upper-bounds the steerability coverage ratio.
Consequently, mutual information enables rollout-free identification of
low-steerability states and serves as a principled proxy for optimizing
steerability in practice.


\section{Hyperparameters}
\label{sec:app:hyperparam}
We provide the hyperparameters used in our training and evaluation setup.
\begin{table}[h]
\centering
\small
\setlength{\tabcolsep}{6pt}
\begin{tabular}{l l}
\hline
\multicolumn{2}{c}{\textbf{Training Configuration}} \\
\hline
Base model & $\pi_{0.5}$ \\
Training steps & 2000 \\
Observation history & 1 \\
Action horizon & 10 \\
Batch size & 128 \\
Finetuning method & Full finetuning \\
Data co-training ratio & LIBERO : SteerGen = 1 : 3 \\
 & SteerGen : SRBC = 1 : 5 \\
Learning rate schedule & Cosine decay \\
Warmup steps & $10{,}000$ \\
Peak learning rate & $5\times10^{-5}$ \\
Decay steps & $1{,}000{,}000$ \\
Final learning rate & $5\times10^{-5}$ \\
\hline
\multicolumn{2}{c}{\textbf{Evaluation Configuration}} \\
\hline
Evaluation timestep range & $[0, 100]$ \\
Evaluation step length & 5 \\
Rollouts per evaluation step & 10 \\
CMI sample size $N$ & 32 \\
\hline
\end{tabular}
\caption{Training and evaluation hyperparameters used in all experiments.}
\label{tab:appendix_hyperparams}
\end{table}

